\documentclass[letterpaper, 10 pt, conference]{ieeeconf}  

\IEEEoverridecommandlockouts                              

\usepackage{amsmath} 

\usepackage{amssymb}  
\usepackage{bbm}
\usepackage{graphicx} 
\usepackage{comment}
\usepackage{booktabs,arydshln}

\makeatletter
\def\adl@drawiv#1#2#3{%
        \hskip.5\tabcolsep
        \xleaders#3{#2.5\@tempdimb #1{1}#2.5\@tempdimb}%
                #2\z@ plus1fil minus1fil\relax
        \hskip.5\tabcolsep}
\newcommand{\cdashlinelr}[1]{%
  \noalign{\vskip\aboverulesep
           \global\let\@dashdrawstore\adl@draw
           \global\let\adl@draw\adl@drawiv}
  \cdashline{#1}
  \noalign{\global\let\adl@draw\@dashdrawstore
           \vskip\belowrulesep}}
\makeatother

\usepackage{threeparttable}
\usepackage{tabularx}
\usepackage{tablefootnote}
\usepackage{tablefootnote}
\usepackage{multirow}
\usepackage[caption=false,font=footnotesize]{subfig}
\usepackage{algorithmic}
\usepackage{algorithm}

\usepackage{color}
\usepackage{xcolor}
\usepackage{colortbl}
\usepackage{listings}
\usepackage[utf8]{inputenc}
\usepackage{threeparttable}
\definecolor{gray}{rgb}{0.7,0.7,0.7}
\usepackage{xspace}
\expandafter\def\expandafter\normalsize\expandafter{%
    \normalsize%
    \setlength\abovedisplayskip{3pt}%
    \setlength\belowdisplayskip{3pt}%
    \setlength\abovedisplayshortskip{3pt}%
    \setlength\belowdisplayshortskip{3pt}%
}

\usepackage[%
	backend=biber,
	url=true,
	block=space,
	maxbibnames=10,
	minbibnames=1,
	bibstyle=ieee,
]{biblatex}
\usepackage[hidelinks]{hyperref}
\AtBeginBibliography{\footnotesize}
\AtEveryBibitem{
	\clearfield{doi}
	\clearfield{isbn}
	\clearfield{issn}
	\clearfield{month}
	\clearfield{publisher}
	\ifentrytype{inproceedings}{
		\clearfield{arxivId}
		\clearfield{archivePrefix}
		\clearfield{eprint}
		\clearfield{url}
		\clearfield{volume}
	}{}
	\ifentrytype{article}{
		\clearfield{arxivId}
		\clearfield{archivePrefix}
		\clearfield{eprint}
		\clearfield{url}
	}{}
	\ifentrytype{report}{
	}{}
}
\renewbibmacro{in:}{%
	\ifentrytype{inproceedings}{%
		\setunit{}
		\addperiod\addspace In \textit{Proc.\ of the}}%
	{\printtext{\bibstring{in}\intitlepunct}}
}
\DeclareSourcemap{
	\maps{
		\map{ 
			\step[fieldsource=url,
				match=\regexp{\{\\\_\}|\{\_\}|\\\_},
				replace=\regexp{\_}]
		}
		\map{ 
			\step[fieldsource=url,
				match=\regexp{\{\$\\sim\$\}|\{\~\}|\$\\sim\$},
				replace=\regexp{\~}]
		}
		\map{ 
			\step[fieldsource=url,
				match=\regexp{\{\\\x{26}\}},
				replace=\regexp{\x{26}}]
		}
	}
}
\title{\LARGE \bf
Sketch-MoMa: Teleoperation for Mobile Manipulator\\ via Interpretation of Hand-Drawn Sketches
}

\author{Kosei Tanada$^{1}$, Yuka Iwanaga$^{1}$, Masayoshi Tsuchinaga$^{1}$, Yuji Nakamura$^{1}$, Takemitsu Mori$^{1}$, \\ Remi Sakai$^{2}$, and Takashi Yamamoto$^{2}$
\thanks{$^{1}$Department of Advanced Robotics Research,
        Toyota Motor Corporation, 543, Kirigahora, Nishihirose-cho, Toyota, Aichi, Japan
        {\tt\small kosei\_tanada@mail.toyota.co.jp}}%
\thanks{$^{2}$Faculty of Information Science, Aichi Institute of Technology,
        1247, Yachigusa, Yakusa-cho, Toyota, Aichi, Japan
        }%
}

\begin{document}

\maketitle
\thispagestyle{empty}
\pagestyle{empty}

\begin{abstract}
    To use assistive robots in everyday life, a remote control system with common devices, such as 2D devices, is helpful to control the robots anytime and anywhere as intended.
    Hand-drawn sketches are one of the intuitive ways to control robots with 2D devices. 
    However, since similar sketches have different intentions from scene to scene, existing work requires additional modalities to set the sketches' semantics.
    This requires complex operations for users and leads to decreasing usability.
    In this paper, we propose Sketch-MoMa, a teleoperation system using user-given hand-drawn sketches as instructions to control a robot.
    We use Vision-Language Models (VLMs) to understand the user-given sketches superimposed on an observation image and infer drawn shapes and low-level tasks of the robot.
    We utilize sketches and the generated shapes for recognition and motion planning of the generated low-level tasks for precise and intuitive operations.
    We validate our approach using state-of-the-art VLMs with 7 tasks and 5 sketch shapes.
    We also demonstrate that our approach effectively specifies more detailed intentions, such as how to grasp and how much to rotate.
    Moreover, we show the competitive usability of our approach compared with the existing 2D interface through a user experiment with 14 participants.
    Our videos and results are available at \textcolor{blue}{\href{https://toyotafrc.github.io/SketchMoMa-Proj}{this link}}.
\end{abstract}

\section{INTRODUCTION}
Social implementation of assistive robots is a crucial solution to labor shortages and improving quality of life.
If a user lives with a robot daily, in addition to autonomous features, easy teleoperation for everyone is helpful to execute tasks or support other family members inside their house and from the outside.
To realize this goal, we have been working on an easy and intuitive teleoperation system enabling users to operate mobile manipulators as intended anytime and anywhere \cite{Hashimoto2013}.

Numerous teleoperation methods have recently been proposed.
In particular, existing work uses VR or MR headsets \cite{cheng2024opentelevisionteleoperationimmersiveactive,iyer2024openteachversatileteleoperation} and wearable gloves \cite{wearablesurvey}, achieving dexterous manipulation with intuitive, easy, and smooth control.
Although users can perform precise control by using these devices as experts do in the laboratory, such devices are not standard in public, making social implementation of the teleoperation systems difficult due to their cost.
Additionally, in exchange for flexibility of operations, users find it difficult to carry these devices, which makes teleoperation less accessible.
From these points of view, we believe that teleoperation using portable 2D devices is more user-friendly.
Existing work proposes versatile teleoperation with 2D devices \cite{stretchcontrolconventional,maniinterfacecomparison}.
However, these methods require continuous interactions and a high memory load on users, increasing workload of teleoperation, especially for mobile manipulation that requires various operations for navigation and manipulation.
It is still challenging for mobile manipulators to achieve easy and intuitive operations with 2D devices as VR and wearable devices can.

\begin{figure}[tb]
    \centering
    \includegraphics[width=1.0\linewidth]{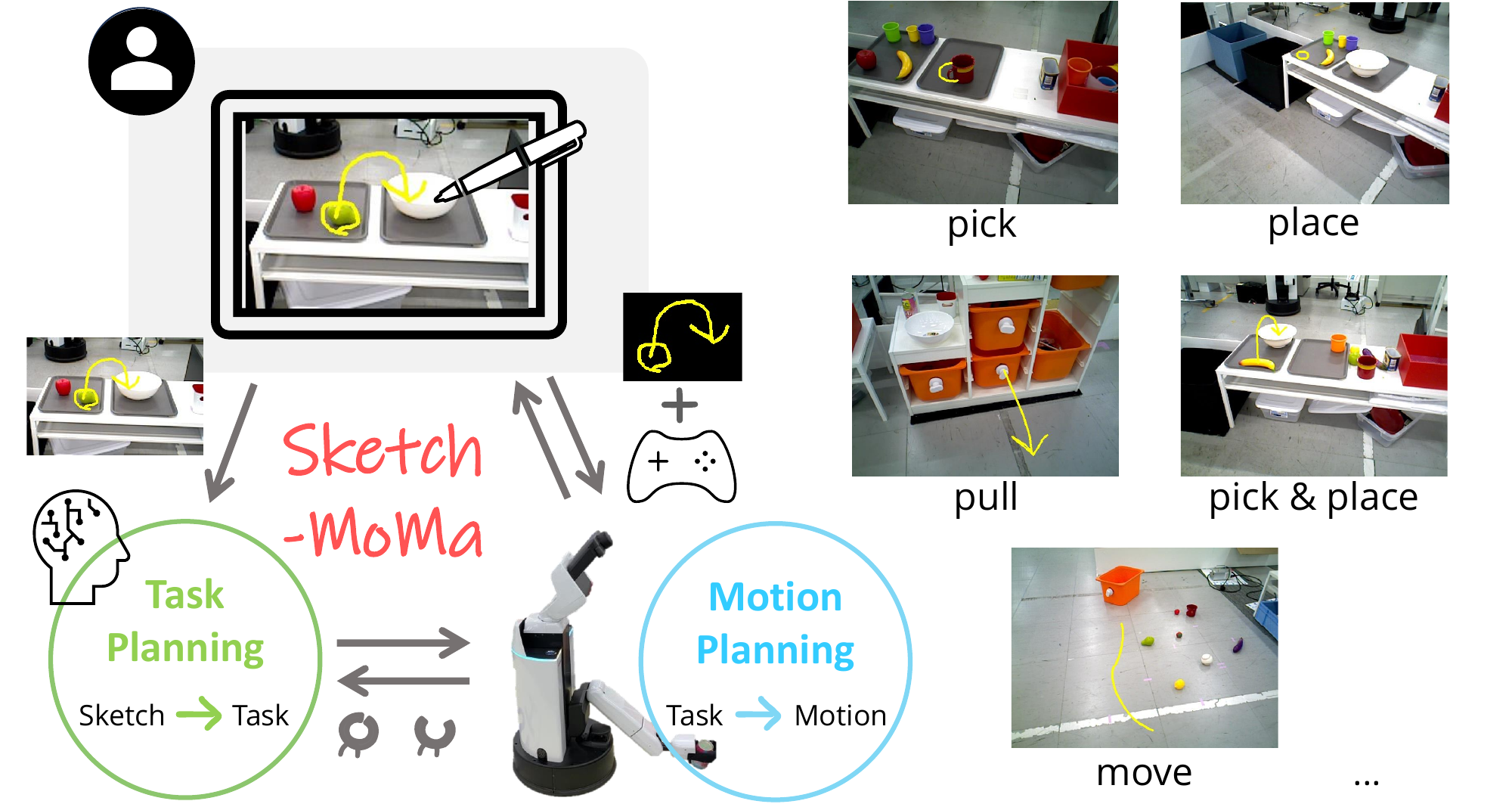}
    \vspace{-15pt}
    \caption{We propose Sketch-MoMa: easy and intuitive teleoperation for mobile manipulation with hand-drawn sketches by combining task planning with a VLM and motion planning.}
    \label{fig:overview}
    \vspace{-10pt}
\end{figure}
Sketches are a user-friendly domain for robot actions \cite{IKnowWhatYouDraw, ASketchInterfaceforRobustandNaturalRobotControl}.
Users can freely communicate with the robot by drawing what they want to do for the robot on canvas (e.g., setting a robot's path and a position to reach).
However, it is necessary to understand what the hand-drawn sketches represent and convert the semantics to the commands for the robot.
Existing work utilizes other modalities \cite{Porfirio_2023,zu2024languagesketchingllmdriveninteractive} or specific shapes \cite{ASketchInterfaceforRobustandNaturalRobotControl} to compensate for the lack of semantics.
Since this requires complex inputs or memorizing all shapes, users may lose the simplicity and intuitiveness of the operations. 
Additionally, even if the tasks require complex user instructions, it is helpful to use unified and intuitive sketches tailored to the task rather than different modalities, such as voice and sketches.
Recent work proposes robotic Foundation Models (FMs) and policies conditioned by hand-drawn goal images \cite{sundaresan2024rtsketchgoalconditionedimitationlearning} or trajectories \cite{gu2023rttrajectory}.
Those approaches have limited intuitiveness as users still depict the instructions in a certain extent of detail.
We believe that the simpler the tasks are, the simpler the required sketches should be, such as circles and arrows, to make the User Interface (UI) easier and more intuitive.
Such simple and intuitive sketches, however, may have ambiguity in grounding to real-world actions.
For example, we need to differentiate similar arrow shapes as shown by two images in the right middle of Fig.~\ref{fig:overview} and transfer them to robot tasks, pick-and-place or pull.
It is challenging to resolve such ambiguity while maintaining the simplicity and intuitiveness of the operations.

In this paper, we propose Sketch-MoMa, a teleoperation system for mobile manipulators that uses a 2D interface and simple hand-drawn sketches.
Users control a mobile manipulator by drawing rough and intuitive sketches on our UI.
We use the capability of Vision-Language Models (VLMs) to interpret the sketches into downstream tasks of the robot and the sketched shapes.
The user-given sketches are utilized in motion planning of the downstream tasks, achieving multiple mobile manipulations with feedback from the operator.
We validate the capability of State-Of-The-Art (SOTA) VLMs with 7 simple tasks and 5 sketch shapes.
We also demonstrate that our approach accurately interprets more complex user intentions by drawing some clues with intuitive shapes and task-related words.
Furthermore, a user study with 14 participants demonstrates the competitive usability of our method compared to a baseline that manually controls each axis of the robot.

\section{RELATED WORK}
\subsection{Teleoperation for Robotics}
Various teleoperation methods using VR/MR \cite{multi-mr,cheng2024opentelevisionteleoperationimmersiveactive,iyer2024openteachversatileteleoperation,dass2024telemomamodularversatileteleoperation}, wearable \cite{wearablesurvey, wang2024dexcapscalableportablemocap}, and customized devices \cite{corrective-shared-autonomy, wu2024gellogenerallowcostintuitive,zhao2023learningfinegrainedbimanualmanipulation} have been proposed.
While these devices enable users to control robots flexibly as intended, they are not commonly used and affordable for end-users, making it difficult to use them for daily teleoperation.
Some approaches focus on portable 2D devices \cite{qin2024anyteleopgeneralvisionbaseddexterous,maniinterfacecomparison, interactive-marker,object_manipulation_with_depth} that are widely used these days.
Cabrera et al. \cite{stretchcontrolconventional} proposed a 2D interface for mobile manipulators that can control each robot axis using buttons on the camera image.
However, the requirements of continuous interactions and a high memory burden increase the users' workload.
Our work applies hand-drawn sketches for the operation with 2D devices.
This provides easy task specifications with simple drawings and reduces required interactions for the operation.

\subsection{Hand-drawn Sketches for Robot Operation}
Sketches have been used for robot control to specify the map \cite{draw-path-and-map, draw-map}, path \cite{Porfirio_2023,sketch-drone, sketch-drone2, sketch-and-run, sketch-path-drawing-for-3d-walkthrough}, trajectory \cite{Zhi2024}, grasping objects \cite{IKnowWhatYouDraw}, constraints \cite{Sun2024}, and tasks \cite{ASketchInterfaceforRobustandNaturalRobotControl}.
Lin et al. \cite{IKnowWhatYouDraw} used object sketches to detect and manipulate the objects.
Porfirio et al. \cite{Porfirio_2023} proposed an end-user system to script robot programs by voice-based interaction and contextualizing the program core with sketches.
While some teleoperation methods focus on manipulation or navigation separately, few approaches consider some mobile manipulation tasks with sketches. 
In addition, due to the lack of the semantics of the sketches, existing work requires other modalities (e.g., voice, texting \cite{Porfirio_2023,zu2024languagesketchingllmdriveninteractive}) or shape detection for each task \cite{ASketchInterfaceforRobustandNaturalRobotControl}).
However, this requires users to input multimodality or remember all shapes for each task, which reduces the simplicity and intuitiveness of the operation. 
Our work presents a teleoperation method for mobile manipulators using sketches and VLMs to interpret the sketches to the desired robot tasks.
This enables users to navigate and manipulate the robot by drawing rough sketches along with each task.

\subsection{Application of Foundation Models for Robotics}
Recently, FMs have achieved significant progress in various fields, and robotics is no exception \cite{hu2023generalpurposerobotsfoundationmodels,firoozi2023foundationmodelsroboticsapplications,kawaharazuka2024realworldrobotapplicationsfoundation,rt22023arxiv,embodimentcollaboration2024openxembodimentroboticlearning}.
Existing work proposes robotic policies \cite{brohan2023rt1roboticstransformerrealworld,rt22023arxiv}, where they take the sensor and state observations with instructions and output robot actions (e.g., end-effector velocity).
Recent work extends these policies and introduces hand-drawn sketches as goal images \cite{sundaresan2024rtsketchgoalconditionedimitationlearning} or trajectory representations with grasping and releasing points \cite{gu2023rttrajectory}.
However, to ensure easy and intuitive teleoperation, we should achieve such robot actions with simpler and more intuitive sketches.
Some approaches apply VLMs and Large Language Models (LLMs) for task and motion planning of the robots.
Existing work grounds the commonsense knowledge of VLMs and LLMs to robot data \cite{dipalo2024keypointactiontokens}, a specific robotic concept \cite{rana2023sayplangroundinglargelanguage}, and code generation \cite{huang2023voxposercomposable3dvalue,liang2023codepolicieslanguagemodel}.
More recently, some methods introduce visual prompting to make VLMs resolve manipulation \cite{liu2024mokaopenvocabularyroboticmanipulation} and navigation tasks \cite{nasiriany2024pivotiterativevisualprompting} in real-world settings.
While much work focuses on text-based instructions with/without visual observation and prompting, few methods propose using instructions given on an image, such as sketches.
If VLMs can interpret roughly drawn instructions on an image into robot tasks directly, the users can operate the robots by just giving simple and intuitive sketches.
However, to the best of our knowledge, no proposed teleoperation methods apply and validate VLMs to understand the sketches as visual instructions for the robots.
We propose a novel teleoperation method using sketches as visual instructions and VLMs to interpret the sketches into robotic tasks.

\begin{figure*}[t]
    \centering
    \includegraphics[width=1.0\linewidth]{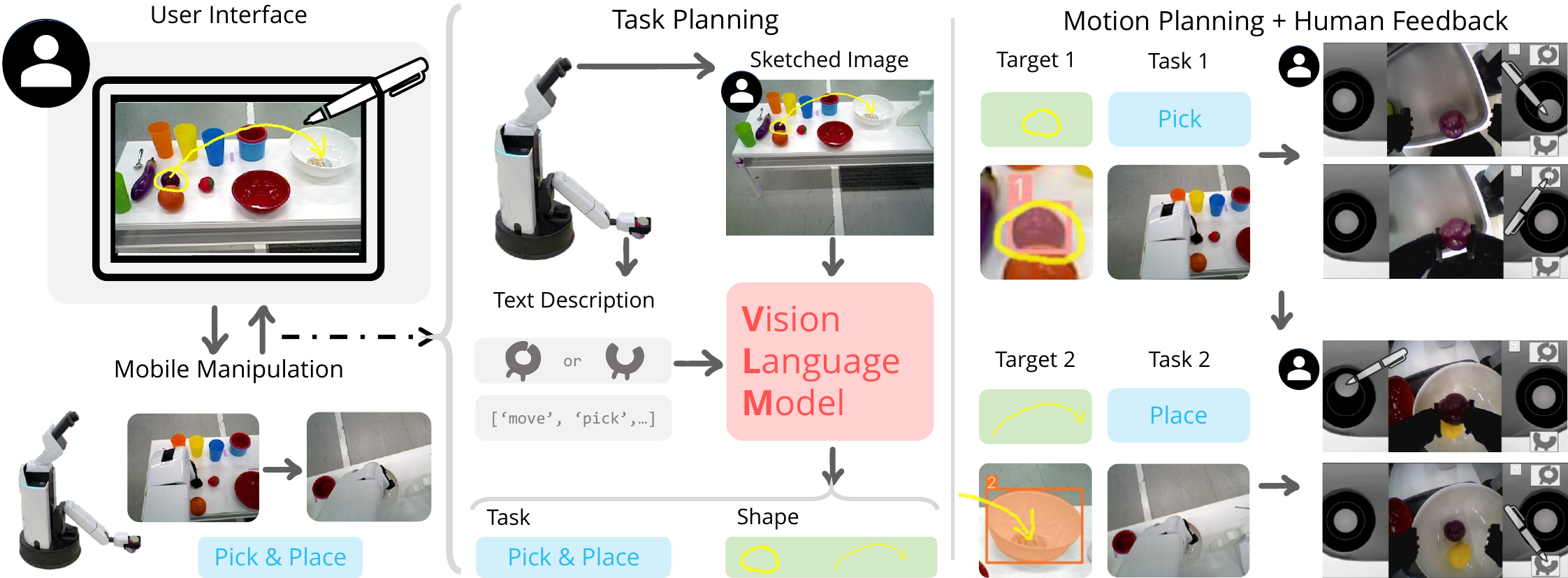}
    \vspace{-18pt}
    \caption{Overview of Sketch-MoMa. Users can interact with a mobile manipulator via a 2D interface by some widgets and drawing sketches on the canvas (left above). We bridge sketch instructions to robot control by implementing task planning with a VLM and motion planning with user-given sketches.}
    \label{fig:system}
    \vspace{-10pt}
\end{figure*}
\section{Methodology}
\subsection{System Overview}
Fig.~\ref{fig:system} shows the system overview of Sketch-MoMa. 
A user draws rough sketches on an observation image from the robot to specify the task. 
A VLM infers indicated tasks based on the given sketches, where the VLM takes text descriptions and the image overlaid by the sketches as inputs and outputs the tasks and the sketched shapes. 
The generated tasks are executed with object recognition and motion planning based on the sketches and the sketched shapes. 
The user provides feedback at specific stages accordingly. 
We describe the task and motion planning in the following sections.

\subsection{Understanding Sketched Instructions with VLMs}
\begin{figure}[tb]
    \centering
    \includegraphics[width=1.0\linewidth]{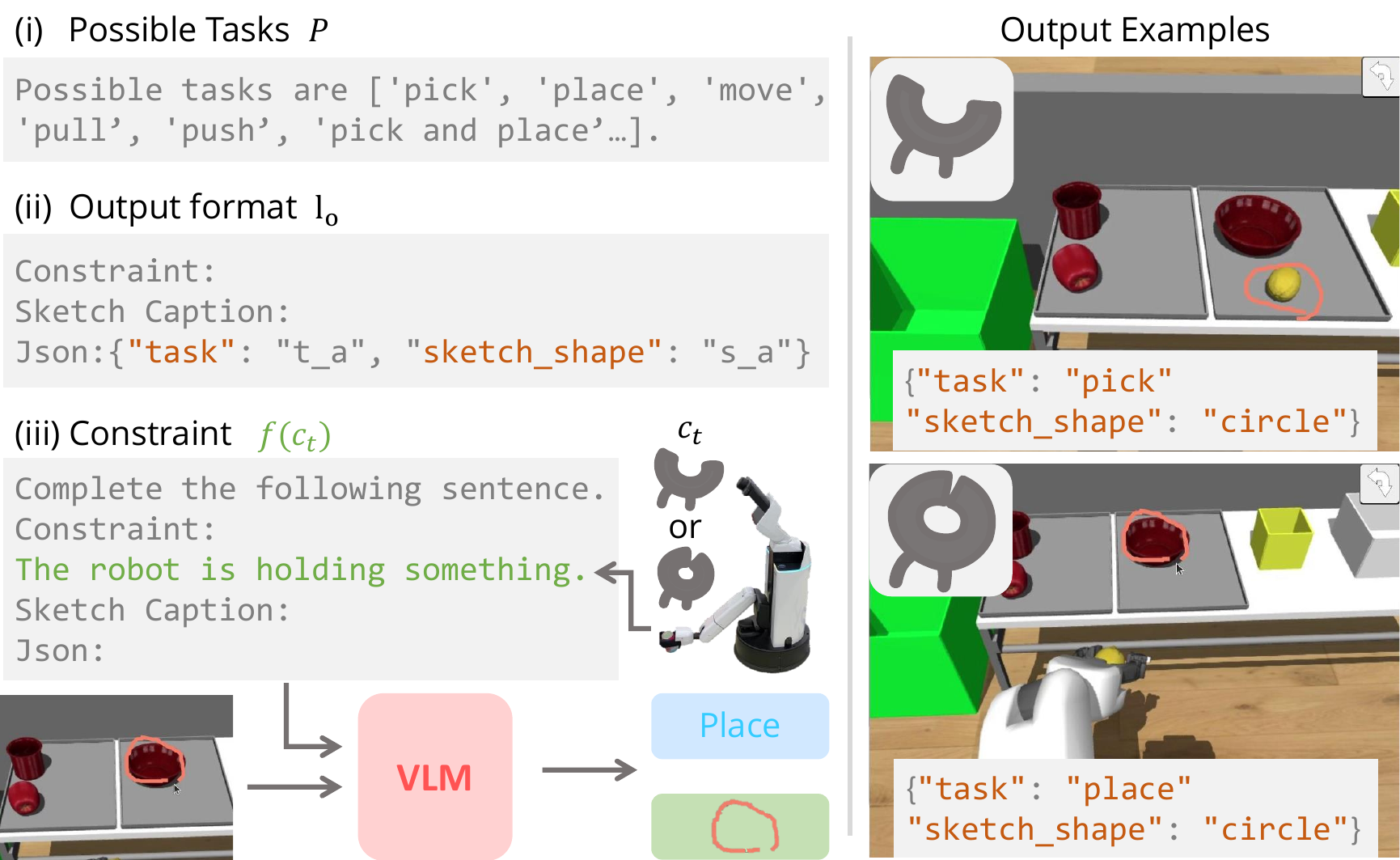}
    \vspace{-18pt}
    \caption{Key text descriptions to ground a VLM to understand sketches.}
    \label{fig:constraint_diff-with_const}
    \vspace{-10pt}
\end{figure}
\subsubsection{Preliminaries}
Generally, a VLM $\mathcal{V}$ takes a sequence of images $\mathcal{I}$ and texts $\mathcal{T}^\mathrm{i}$ and outputs texts $\mathcal{T}^\mathrm{o}$.
$$
\label{eq:normal-vlm}
\mathcal{T}^\mathrm{o} = \mathcal{V}(\mathcal{I}, \mathcal{T}^\mathrm{i}). \eqno{(1)}
$$
In our method, $\mathcal{I} = I(o_t, S)$, where $o_t$ is an observation image at time $t$, $S$ is sketches drawn by a user, and $I$ is a function to overlay $S$ on $o_t$.
$\mathcal{T}^i$ is composed of two parts: pre-defined $\mathrm{l_p}$ and dynamically composed $l_\mathrm{d}$. 
Here, $\mathcal{T}^i = [\mathrm{l_p}, l_\mathrm{d}]$.
$\mathrm{l_p}$ is defined as $\mathrm{l_p} = [\mathrm{l_s}, \mathrm{l_e}(\mathrm{l_o}), \mathrm{l_q}]$, where $\mathrm{l_s}$ describes the system overview including a list of possible tasks $P$, $\mathrm{l_e}$ is examples of answers following a text output format $\mathrm{l_o}$, and $\mathrm{l_q}$ asks a request to the VLM.
$l_\mathrm{d}$ is defined as $l_\mathrm{d} = T(f(c_t), \mathrm{l_o})$, where $c_t$ is a constraint of the robot at time $t$, $f$ is a mapping function from $c_t$ to text, and $T$ is a text-composing function.
$\mathrm{l_o}$ has the same format provided in $\mathrm{l_e}$.

\subsubsection{Grounding VLMs for Sketch Instructions}
\label{subsec:prompt}
We introduce three key prompts to understand hand-drawn instructions. Some parts of the actual prompts are shown in Fig.~\ref{fig:constraint_diff-with_const} left. Our prompt is available at \href{https://toyotafrc.github.io/SketchMoMa-Proj}{our website}.

\noindent
(i) \lstinline{Possible Tasks} $P$ provides a list of tasks that can be executed. This enables VLMs to select a task according to the given sketches.

\noindent
(ii) \lstinline{Output Format} $\mathrm{l_o}$ provides a format to generate downstream tasks and sketched shapes. The format takes the form of JSON style and includes \lstinline{task} and \lstinline{sketch_shape} keys. The value of \lstinline{task} is executed directly as a Python function, and \lstinline{sketch_shape} is provided as the argument in the low-level module.

\noindent
(iii) \lstinline{Constraint} $f(c_t)$ describes the current constraint of the robot. Imagine that a user draws a circle on a bowl, as shown in Fig.~\ref{fig:constraint_diff-with_const} right. If there are the circle and task options \lstinline{pick} and \lstinline{place}, it is unclear which task the user intended to instruct. To resolve this ambiguity, we add a description of current \lstinline{Constraint}. Specifically, we recognize the gripper state and describe whether the robot is holding an object or not in text format (e.g., ``The robot is holding an object"). Incorporating \lstinline{Constraint} prompt enables VLMs to infer tasks considering the current state of the robot.

\noindent
We use VLMs for our task planning that can be handled by some machine learning approaches since versatile common sense knowledge of VLMs can be robustly utilized in the interpretation of hand-drawn sketches in a few-shot manner. This enables easy application to new tasks and shapes by appending them in $P$ and $\mathrm{l_e}$. 

\subsection{Motion Planning with User-given Sketches.}
\begin{figure}[tb]
    \centering
    \includegraphics[width=1.0\linewidth]{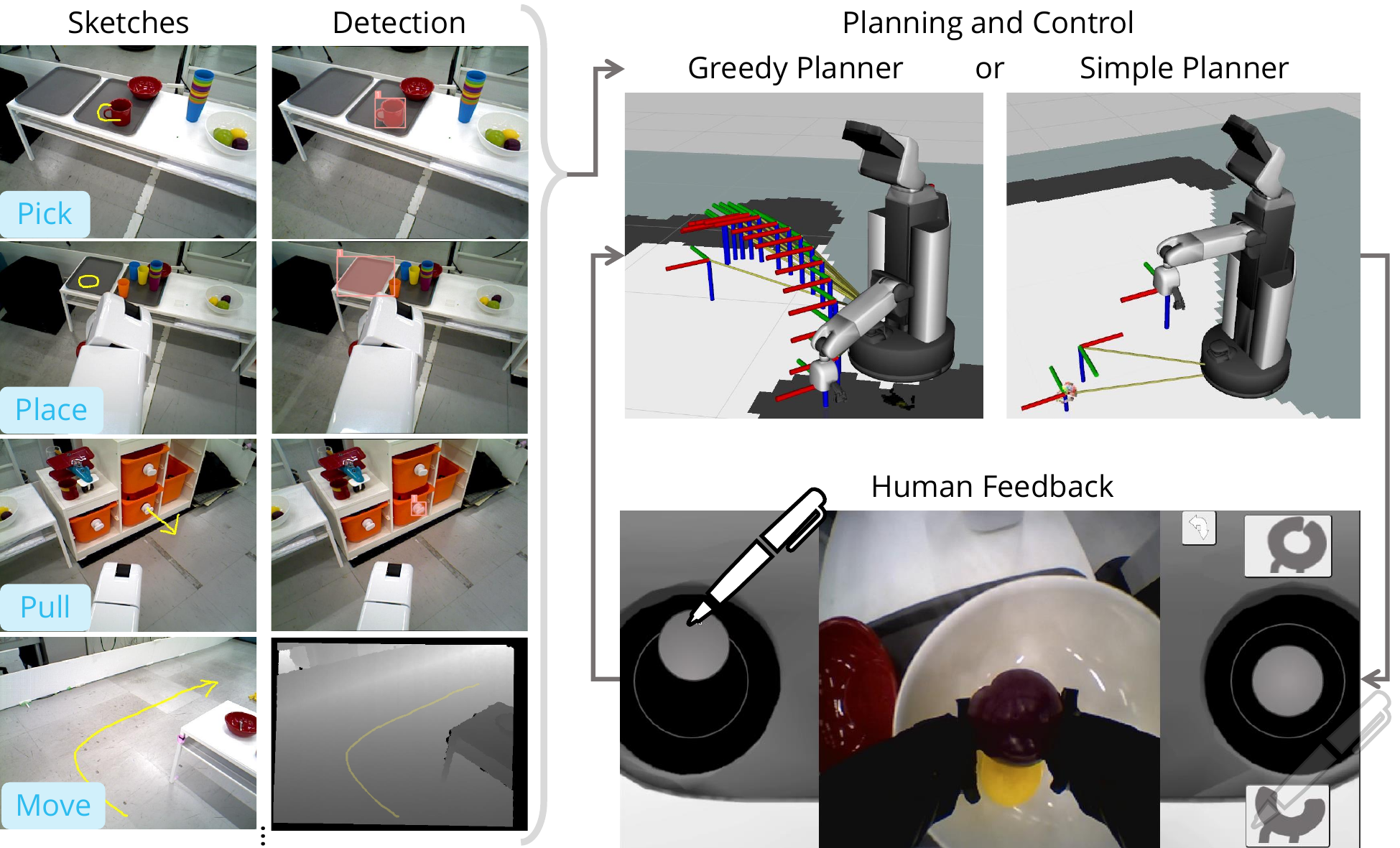}
    \vspace{-18pt}
    \caption{In motion planning, we detect objects and their poses based on tasks, given sketches, and their shapes. We then plan the end-effector trajectory to the target objects. The robot asks for feedback from the user after reaching the target objects.}
    \label{fig:motion_planning}
    \vspace{-10pt}
\end{figure}
We describe how we use user-given sketches and inferred sketch shapes in downstream modules. We also describe the features of our human feedback system.
\subsubsection{Manipulation}
\label{sec:object_detection}
\textbf{Object Detection:} We introduce object segment detection based on the sketches.
We first detect all object segmentation in the scene and then extract the largest object area in bounding boxes constructed by the sketches.
We use minimum and maximum pixel positions of sketches to form the bounding boxes.
When the inferred sketch shape is ``arrow", we use the sketch's starting and ending points to create the bounding boxes.
We use FastSAM \cite{zhao2023fastsegment} and track the detected segmentation with XMem++ \cite{bekuzarov2023xmemproductionlevelvideosegmentation}.

\noindent
\textbf{Grasp Pose Selection:} We introduce grasp pose detection based on the generated sketch shape.
When users draw a circle or arrow shape, we use object pose detection based on the tracked segmentation described in Sec.~\ref{sec:object_detection}.
We use Scaled-Balanced-Grasp \cite{ma2022scalebalanced6dofgrasp} to detect the grasp pose.
When the user draws a U-shape, we use a heuristic approach that calculates the reaching direction from the right, left, front, or above of the object. 
Examples of the U-shape sketches are shown in Fig.~\ref{fig:fig_variation1}.

\noindent
\textbf{Trajectory Planning and Whole-body Control:}
The robot plans the end-effector trajectory from the current pose to the detected target pose and performs the whole-body control. We use a static greedy planner of the end-effector trajectory inspired by Voxposer \cite{huang2023voxposercomposable3dvalue} and the whole-body control deployed in HSR \cite{hsr-journal}.
In our user study (Sec.~\ref{subsec:ux-setup}), we use a simple planner that approaches the target step by step for time efficiency.

\noindent
\textbf{Human Feedback:}
In teleoperation, users may intend to make slight adjustments to the robot's motion.
Although more accurate estimation reduces this needs, we believe that such feedback functions have a crucial role for usability. 
We provide a human feedback system to adjust the position that the robot has reached. After reaching a target object, the robot asks the operator to adjust the end-effector position and grasp or release the object using the joysticks and grasp/release buttons (Fig.~\ref{fig:motion_planning} right bottom). 
The left joystick controls the end-effector in depth direction, and the right joystick supports translation in the image plane of the hand camera. 
This manual operation is limited to fine adjustments, minimizing the operation burden while maintaining the intuitiveness for users.

\subsubsection{Navigation}
\textbf{Path Specification:} We use the sketches to set the path of the robot on the floor by extracting the depth values at the sketch points (Fig.~\ref{fig:motion_planning} left bottom).

\noindent
\textbf{Joystick Control:} We also provide robot base control to adjust the base position with joysticks for human feedback. The operator uses a left joystick for translation of the base and a right joystick for view control. 
\section{Experiments}
\label{sec:experiment}
\begin{table}[tb]
    \caption{Task-and-Shape Pairs included in Dataset}
    \vspace{-15pt}
    \label{tab:task-shape-pair}
    \begin{center}
    \begin{tabular}{cccccc}
        \toprule
        & Circle & U-shape & Arrow & Path & Circle\&Arrow \\
        \midrule
        Pick & \checkmark & \checkmark &  &  &  \\
        Place & \checkmark &  & \checkmark &  &  \\
        Move &  &  & \checkmark & \checkmark  & \\
        Pull & \checkmark & \checkmark & \checkmark &  & \checkmark  \\
        Push &  &  & \checkmark &  & \checkmark \\
        Drop &  &  & \checkmark &  &  \\
        Pick\&Place &  &  & \checkmark &  & \checkmark \\
        \bottomrule
    \end{tabular}
    \end{center}
    \vspace{-20pt}
\end{table}
\begin{figure}[tb]
    \centering
    \subfloat[Pick\label{fig:fig_variation1}]{
    \includegraphics[height=0.16\linewidth]{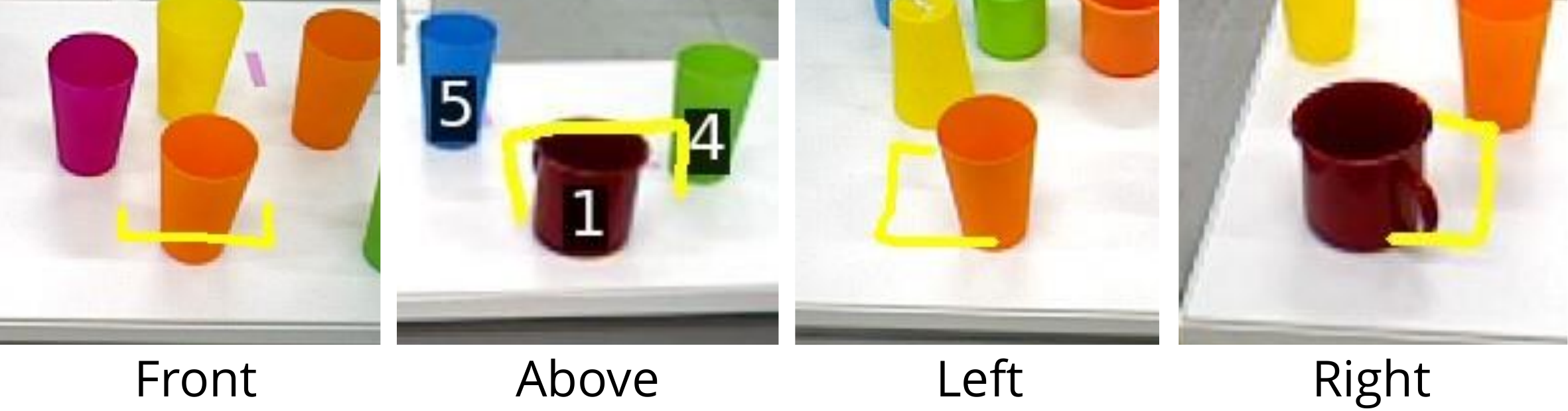}}
    \subfloat[Rotate\label{fig:fig_variation2}]{\includegraphics[height=0.16\linewidth]{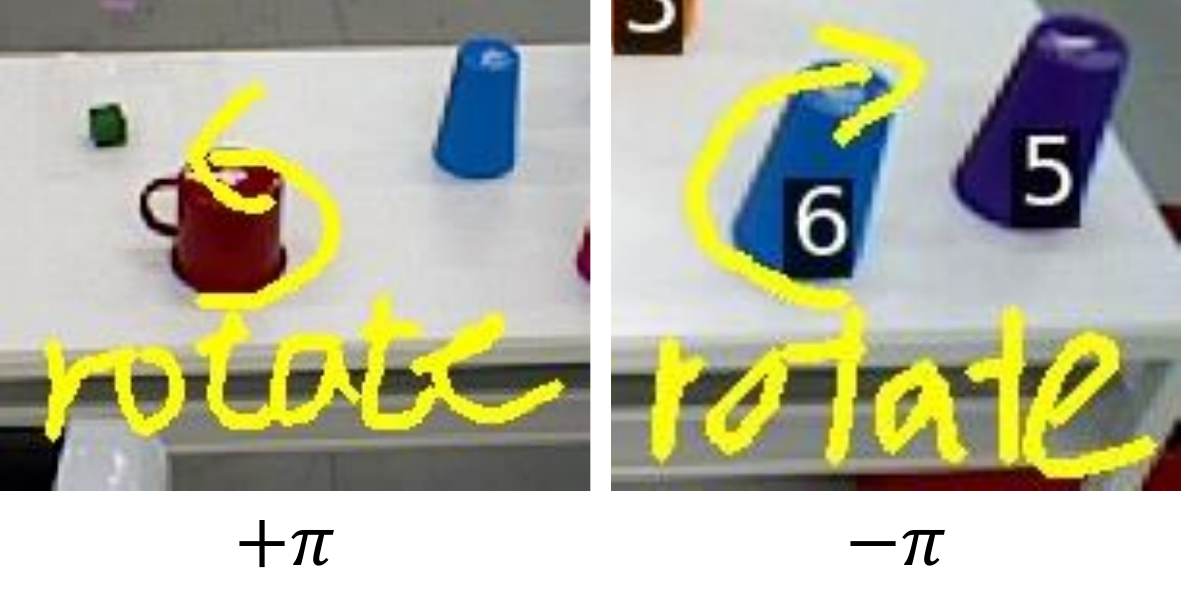}}
    \vspace{-4pt}
    \caption{Variation of sketches for the instruction of detailed motions. We provide 4 directions for grasping with a U-shape and two rotations with an arrow shape. We set numbers on objects with SoM \cite{yang2023setofmarkpromptingunleashesextraordinary} for VoxPoser \cite{huang2023voxposercomposable3dvalue} to make VLMs understand the specified objects visually.}
    \label{fig:fig_variation}
    \vspace{-12pt}
\end{figure}
\begin{figure}[t]
    \centering
    \label{fig:ux_details}
    \subfloat[Environment\label{fig:ux_env}]{
		\includegraphics[height=0.22\linewidth]{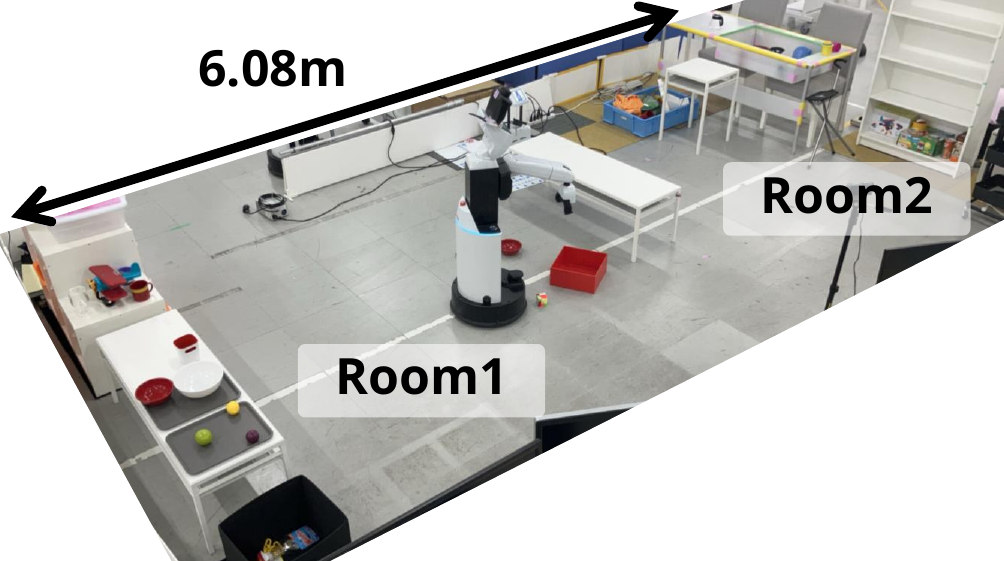}}\hfill
    \subfloat[Task\label{fig:task_in_ux}]{
		\includegraphics[height=0.22\linewidth]{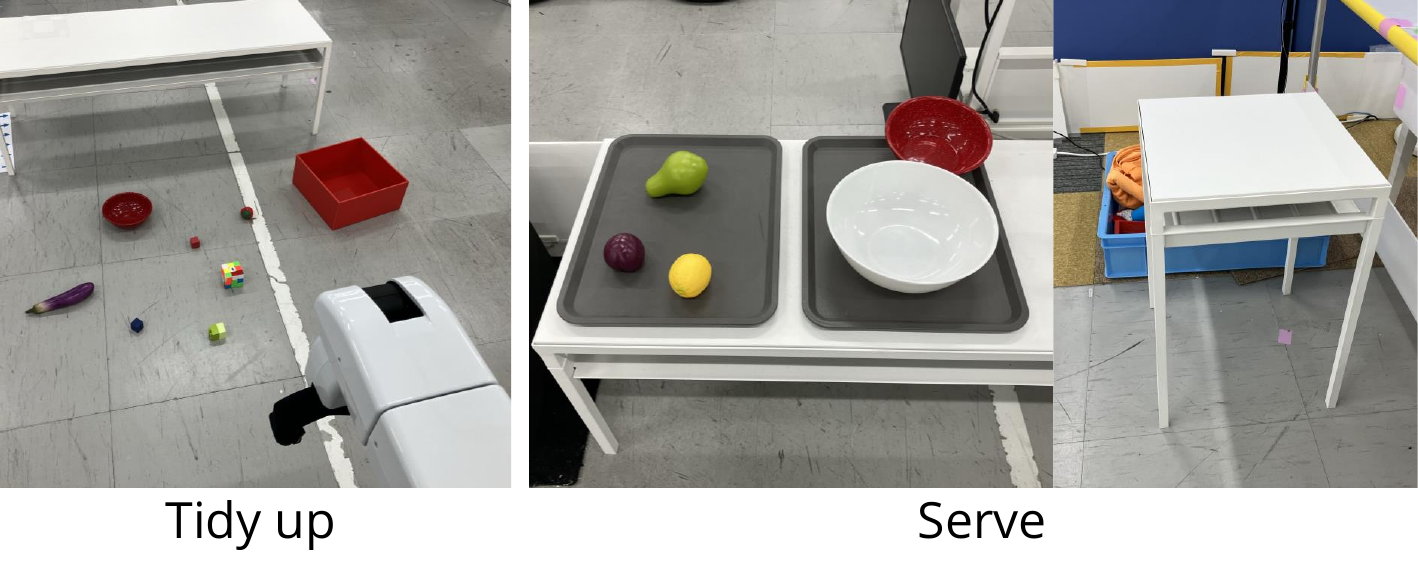}}
    \vspace{-4pt}
    \caption{Environment and task scenes of our user study.}
    \vspace{-15pt}
\end{figure}
We consider the following three research questions.

\noindent
\textbf{Q1:} Can SOTA VLMs interpret hand-drawn sketches on an observation image to robot tasks with our proposal?

\noindent
\textbf{Q2:} Does the proposed method interprets more complicated intentions with specific clues, such as shapes and words?

\noindent
\textbf{Q3:} Does the proposed teleoperation have positive user experiences in mobile manipulation?
\subsection{Inference Accuracy of SOTA VLMs}
\label{subsec:rq1-setup}
We validate the capability of VLMs to generate shapes and robot tasks from hand-drawn sketches.
We collect a dataset containing 5000 pairs of sketched images and texts, including 7 tasks and 5 sketch shapes.
Each image overlaid by sketches captures a scene that is similar to the World Robot Summit (WRS) competition site \cite{wrs}.
The texts are the ground truth of the tasks and shapes.
The 7 tasks are 6 single tasks (Pick, Place, Move, Pull, Push, and Drop) and a multi-task, Pick-and-Place (P\&P).
The 5 sketch shapes are Circle, U-shape, Arrow, Path, and Circle-and-Arrow (C\&A).
Table~\ref{tab:task-shape-pair} shows the pairs of tasks and sketch shapes included in the dataset.
We provide prompts described in Sec.~\ref{subsec:prompt} in a zero- (without examples) and few-shot (with examples) manner.
We randomly select 50 pairs for each task and shape and show \textit{inference success rate} (ISR) with them.
ISR is set as a success when the VLM reasons for the correct task or shape.
We validate two models: GPT-4V \cite{gpt4v} and Gemini Pro \cite{gemini}.

Table~\ref{tab:inference_task_success} left shows the validation results of task inference.
Gemini gets higher ISRs on 5 tasks, and GPT-4V (GPT) scores the highest on Pick.
Both models score higher in a few-shot manner than a zero-shot manner, except for Push.
All ISRs on Push are low in the same way.
Table~\ref{tab:inference_task_success} right shows the results of shape inference.
Gemini scores the highest on all shapes except for Arrow in a zero-shot manner, while GPT gets a higher score on Arrow in a few-shot manner.
GPT scores higher in a few-shot manner on U-shape, Arrow, and Path than in a zero-shot manner.
Gemini gets higher scores in a zero-shot manner for most tasks than a few-shot manner.
\begin{table*}[tb]
    \begin{center}
    \begin{minipage}{0.7\linewidth}
    \caption{Inference Success Rates [\%] with 50 pairs on our dataset.}
    \vspace{-10pt}
    \label{tab:inference_task_success}
    \begin{threeparttable}
        \begin{tabular}{ccccc|ccccc}
            \toprule
            \multirow{2}{*}{\textbf{Task}} & \multicolumn{2}{c}{Zero} & \multicolumn{2}{c|}{Few} & \multirow{2}{*}{\textbf{Shape}} & \multicolumn{2}{c}{Zero} & \multicolumn{2}{c}{Few} \\
                \cmidrule(r){2-3}  \cmidrule(r){4-5} \cmidrule(r){7-8}  \cmidrule(r){9-10}
            & GPT & Gemini & GPT & Gemini &  & GPT & Gemini & GPT & Gemini \\
            \midrule
            Pick & 46 & 46 & \textbf{84}\tnote{*} & 52 & Circle & 72 & \textbf{98} & 64 & 72 \\
            Place & 78 & 78 & 94 & \textbf{96} & U-shape & 24 & \textbf{88} & 58 & 24 \\
            Move &  96 & 96 & 84 & \textbf{100} & Arrow & 70 & 70 & \textbf{84} & 70 \\
            Pull & 50 & 50 & 80 & \textbf{84} & Path &  74 & \textbf{100} & 90 & 74 \\
            Push & \textbf{14} & \textbf{14} & 12 & 4 & Circle\&Arrow & \textbf{20} & \textbf{20} & 18 & \textbf{20} \\
            Drop & 2 & 2 & 50 & \textbf{74} & - & - & - & - & - \\
            Pick\&Place & 36 & 36 & 58 & \textbf{60} & - & - & - & - & - \\
            \bottomrule
        \end{tabular}
        \begin{tablenotes}[para,flushleft,online,normal]
            \item[*] Bold style shows the highest score in each item.
        \end{tablenotes}
        \vspace{-18pt}
    \end{threeparttable}
    \end{minipage}
    \hfill
    \begin{minipage}{0.25\linewidth}
    \centering
    \vspace{-6pt}
    \caption{Task Validation in Real World.}
    \vspace{-18pt}
    \label{tab:task_validation}
    \begin{threeparttable}
        \hfill
        \begin{tabular*}{\linewidth}{c@{\extracolsep{\fill}}cc}
            \toprule
            \textbf{Task} & ISR\tnote{1} & TSR\tnote{2} \\
            \midrule
            Pick & 10/10 & 7/10 \\
            Place & 10/10 & 7/10 \\
            Move & 10/10 & 9/10 \\
            Pull & 5/10 & 4/10 \\
            Push & 0/10 & 0/10 \\
            Pick\&Place & 10/10 & 7/10 \\
            \bottomrule
            & & \\
        \end{tabular*}
        \begin{tablenotes}[para,flushleft,online,normal]
            \item[1] \textit{inference success rate}
            \item[2] \textit{task success rate}
        \end{tablenotes}
        \vspace{-18pt}
    \end{threeparttable}
    \end{minipage}
    \end{center}
\end{table*}
\subsection{Task Reliability in Real World.}
We validate the same 6 tasks as Sec.~\ref{subsec:rq1-setup} except for Drop in the real world.
We draw Circle and U-shape on Pick, Circle on Place, Arrow on Pull and Push, C\&A on P\&P, and Path on Move.
We use our feedback system that is operated by an expert.
We test each task 10 times in the WRS environment and show ISRs and \textit{task success rates} (TSRs).
TSR is successful when the task is executed as desired.
We use GPT-4V as a VLM in a few-shot manner.

Table~\ref{tab:task_validation} shows ISRs and TSRs in a real-world environment.
We demonstrate high ISRs and TSRs on Pick, Place, Move, and Pick\&Place. ISRs are lower on Pull and Push than the other tasks, especially on Push.

\subsection{Robot Operation with Sketches for Detailed Motions}
\begin{table}[tb]
    \centering
    \begin{threeparttable}
        \caption{Validation with Sketches Containing Detailed Specification.}
        \label{tab:detailed_intended_sketches}
        \begin{tabular*}{\linewidth}{cccccccc}
            \toprule
            \multirow{2}{*}{\textbf{Task}}& \multirow{2}{*}{\textbf{Variation}}  & \multicolumn{2}{c}{Ours} & \multicolumn{2}{c}{CaP} \cite{liang2023codepolicieslanguagemodel} & \multicolumn{2}{c}{VoxPoser \cite{huang2023voxposercomposable3dvalue}} \\
            \cmidrule(r){3-4} \cmidrule(r){5-6} \cmidrule(r){7-8}
            & & VSR\tnote{1} & TSR\tnote{2} & VSR & TSR & VSR & TSR \\
            \midrule
            Pick & Right & \textbf{9/10}\tnote{3} & 7/10 & 0/10 & 0/10 & 4/10 & 0/10 \\
                 & Left  & \textbf{6/10} & 5/10 & 2/10 & 1/10 & 4/10 & 0/10 \\
                 & Above & \textbf{7/10} & 7/10 & 1/10 & 1/10 & \textbf{7/10} & 0/10 \\
                 & Front & \textbf{8/10} & 7/10 & 2/10 & 1/10 & 6/10 & 0/10 \\
            \midrule
            Rotate & $+\pi$ & \textbf{9/10} & 8/10 & 3/10 & 1/10 & 5/10 & 4/10 \\
                   & $-\pi$ & \textbf{9/10} & 9/10 & 1/10 & 0/10 & 3/10 & 2/10 \\
            \bottomrule
        \end{tabular*}
        \begin{tablenotes}[para,flushleft,online,normal]
            \item[1] \textit{variation success rate}
            \item[2] \textit{task success rate}
            \item[3] Bold style shows the highest score in VSRs of each item.
        \end{tablenotes}
        \vspace{-5pt}
    \end{threeparttable}
\end{table}
\begin{table}[t]
    \centering
    \caption{Statistical Result of Metrics on User Study.}
    \label{tab:ux-metrics}
    \vspace{-10pt}
    \begin{threeparttable}
        \centering
        \begin{tabular*}{\linewidth}{c@{\extracolsep{\fill}}cccccc}
            \toprule
            \multirow{2}{*}{\textbf{Metrics}} & \multicolumn{2}{c}{Ours} & \multicolumn{2}{c}{Baseline} & & \\
            \cmidrule(r){2-3} \cmidrule(r){4-5}
            & $\mu$\tnote{*} & $s$\tnote{*} & $\mu$ & $s$ & $p$\tnote{*} & $g$\tnote{*} \\
            \midrule
            $m_\mathrm{w}$ & \textbf{45.2}\tnote{\dag} & 19.0 & 53.4 & 17.0 & 0.09 & 0.45 \\
            $m_\mathrm{d}$ & \textbf{2.93} & 1.00 & 2.50 & 1.09 & 0.16 & 0.41 \\
            $m_\mathrm{i}$ & 2.71 & 0.99 & \textbf{2.35} & 1.01 & 0.13 & -0.36 \\
            $m_\mathrm{e}$ & \textbf{1.50} & 0.51 & 1.64 & 0.49 & 0.34 & 0.28 \\
            \bottomrule
        \end{tabular*}
        \begin{tablenotes}[para,flushleft,online,normal]
            \item[*] $\mu$: mean, $s$: corrected sample standard deviation, $p$: p value, $g$: Hedge's $g$
            \item[\dag] Bold style shows higher scores.
        \end{tablenotes}
    \end{threeparttable}
    \vspace{-10pt}
\end{table}
We test the adaptability of our method on tasks including more complicated intentions.
As validation tasks, we select Pick as a verified task and Rotate as a new task.
We provide some clues to instruct detailed intentions for each task.
Fig.~\ref{fig:fig_variation} shows the variation of sketches for each task.
On Pick, we use U-shape sketches from 4 directions: right, left, above, and front, instructing how to grasp it.
On Rotate, we use $+\pi$ and $-\pi$ Arrow and the word ``rotate'' sketches, instructing task, how, and how much to rotate the object.
We set VoxPoser \cite{huang2023voxposercomposable3dvalue} and CodeasPolicies (CaP) \cite{liang2023codepolicieslanguagemodel} as baselines.
These two methods achieve various robotic tasks by code generation with LLMs and VLMs.
We introduce the baselines since we consider the possibility to generate function parameters along the sketches as desired (e.g., \lstinline{grasp_from("right")} or \lstinline{rotate(}$\pi$\lstinline{)}).
We design text prompts for baselines with direction (right, left,  above, and front) and rotation ($\pi/2$, $-\pi/2$, $\pi$, $-\pi$) parameters.
For baselines, we use a VLM for the initial code generation and a LLM for the following generations.
For VoxPoser, since we do not use text instructions to specify target objects, we use SoM \cite{yang2023setofmarkpromptingunleashesextraordinary} to overlay numbers on the objects and provide the list of objects' numbers to the VLM and LLM.
We use \textit{variation success rate} (VSR) and TSR as metrics.
VSR is set as a success for baselines if the VLM reasons the correct direction/rotation parameters (Cap and VoxPoser) and object (VoxPoser).
VSR is successful for our method if the VLM outputs the correct sketch shapes and the parameters are calculated as desired.
We set TSR as a success when the robot completes the desired task.
We use GPT-4V as a VLM since ISR is higher on Pick, as shown in Table~\ref{tab:inference_task_success}.
We also use GPT-4 \cite{gpt4} as an LLM because of the compatibility of implementation with GPT-4V.

Table~\ref{tab:detailed_intended_sketches} shows VSRs and TSRs on each task.
Our method demonstrates high VSRs and TSRs on both tasks, while baselines mostly fail to generate the correct parameters of the motions.

\subsection{User Study of Teleoperation}
\label{subsec:ux-setup}
In our user experiment\footnote{The experiment is approved by Toyota Motor Corporation's Research Ethic Review Board (ID: 2024TMC209).},
$N=14$ participants from our internal workers join the experiment (12 men and two women, $M_{\mathrm{age}}=39.6$).
The participants control HSR \cite{hsr-journal} in the WRS environment (Fig.~\ref{fig:ux_env}) at most 4 hours, two hours with our method and the other two hours with a baseline \cite{stretchcontrolconventional} that controls each axis of the robot with a 2D interface.
After the participants approve an informed consent, they practice controlling HSR until they are used to it.
Then, they challenge two tasks: Tidy-up and Serve (Fig.~\ref{fig:task_in_ux}). On Tidy-up, they clear away 7 objects on the floor into a red box. On Serve, they move three fruit toys on a tray into a white bowl in Room 1 and carry the bowl to a table in Room 2.
They have 20 minutes for each task.
We explain to the participants that they should perform the tasks carefully, not quickly.
After the experiments, they answer 4 questionnaires for the evaluation.
The one is NASA-TLX \cite{NASA-TLX}. NASA-TLX first asks participants to score workload with 6 subjective: Mental Demand, Physical Demand, Temporal Demand, Performance, Effort, and Frustration, from 0 to 100.
Then, it asks which subjective the participants think is more important for all pairs of 6 subjective (15 pairs) to get an individual weight for the subjective.
The other three questions are about difficulty, intuitiveness, and enjoyment of the operation in 5 Likert scales \cite{Likert1932}.
We statistically analyze 4 metrics: Weighted Workload $m_\mathrm{w}$ of NASA-TLX, difficulty $m_\mathrm{d}$, intuitiveness $m_\mathrm{i}$, and enjoyment $m_\mathrm{e}$.
We also measure the means of execution time, $t_\mathrm{tidy}$ and $t_\mathrm{serve}$, and success rates, $r_\mathrm{tidy}$ and $r_\mathrm{serve}$.
We use GPT-4V since it scores reliable ISRs on Pick, Place, and Move, as shown in Table~\ref{tab:inference_task_success}.

Table~\ref{tab:ux-metrics} and Table~\ref{tab:ux-performance} show results, means $\mu$, corrected sample standard deviations $s$, p values $p$, and Hedge's $g$ of the user experiment.
While our method scores longer time and low success rates in both tasks, it shows competitive usability compared with a baseline.
Our method scores lower $m_\mathrm{w}$ and better $m_\mathrm{d}$ and $m_\mathrm{e}$.
Although our $m_\mathrm{i}$ is still less than the baseline, half and more participants answer above ``Neutral''.
We observe small effect sizes ($|g|>0.2$) in all metrics.
We discuss these results in Sec.~\ref{subsubsec:ux-disscussion}.

\subsection{Discussion}
\subsubsection{Q1. Can SOTA VLMs interpret hand-drawn sketches on an observation image to downstream robot tasks?}

\noindent
\textbf{Task Inference:} High ISRs with GPT and Gemini, except for Push, show that SOTA VLMs can understand sketched instructions and select correct tasks with our framework.
Higher ISRs in most tasks in a few-shot manner than a zero-shot manner show that we can ground the VLMs in our concept.
We score fewer ISRs on Pull since the VLM interprets the arrow sketches for Pull to Move.
The instruction difference between Move and Pull depends on visual background on images, making it difficult to differentiate them.
We also score a significantly fewer TSR on Push since the VLM interprets the given arrow sketches to Pull or Move.
The difference in instructions among these tasks depends on visual background and arrow direction. We consider that the VLMs cannot yet interpret the spatial semantics between the scene and sketches since they do not identify the arrow's directional semantics on the scene.

\noindent
\textbf{Shape Inference:}
Gemini observes higher ISRs in a zero-shot manner on Circle, Arrow, U-shape, and Path, while GPT observes a higher ISR on Arrow in a few-shot manner.
This shows that we can utilize the VLM's capability to detect the differences of some simple sketches.
We observe lower ISRs for Circle\&Arrow, demonstrating difficulties for VLMs in recognizing them separately.

\noindent
\textbf{Task in Real World:}
High TSRs in the real world, except for Pull and Push, indicate that our approach enables users to specify certain tasks using simple sketches. TSRs are relatively low for Pull and Push for the same reasons discussed regarding task inference accuracy.

\subsubsection{Q2. Does the proposed method achieve tasks including more complicated intentions with specific clues?}
Our approach effectively realizes more complex motions with sketches.
A VLM tends to reason wrong parameters with CaP on both tasks.
Although the VLM improves the accuracy with VoxPoser, it is still less successful than our approach and has more difficulty in generating downstream codes, leading to significantly low TSRs.
This shows that it is difficult for the VLM to reason the motion parameters from sketched instructions, and our approach effectively connects the sketches to robot operation.
\subsubsection{Q3. Does the proposed method have positive user experiences in mobile manipulation?}
\label{subsubsec:ux-disscussion}
Our method scores competitive usability to the baseline.
Despite taking longer $t_\mathrm{tidy}$ and $t_\mathrm{serve}$ and scoring lower $r_\mathrm{tidy}$ and $r_\mathrm{serve}$ with our method, which may lead to higher Effort and Frustration, they feel more positive about $m_\mathrm{w}$ with our method than a baseline.
They also feel positive in $m_\mathrm{d}$ and $m_\mathrm{e}$.
Additionally, while our method gets lower $m_\mathrm{i}$ than the baseline, half and more participants score better than ``Neutral''.
These results demonstrate that our method has the potential to be an intuitive and easy interface for mobile manipulation.
Since user experiences such as $m_\mathrm{i}$ and $m_\mathrm{w}$, especially for Performance, Effort, and Frustration, can be improved once the autonomous features are stabilized, we need to conduct broader user experiments after the system improvement in the future.
In fact, four of 6 participants who score ``Neutral" and ``False" for $m_\mathrm{i}$ experience inaccurate autonomous movement, and the other two complain of incorrect object detection.
Some participants also feel irritated about waiting for the answer from the VLM.
Introducing more robust autonomous features and deploying faster inference can improve the usability.
Most participants complete Tidy-up with our method, while half do not finish Serve accurately.
In failure cases on Serve, 4 participants drop objects while carrying the bowl, and three collide with the environment in navigation.
In addition to stable object handling, it is essential to provide users with a sufficient view of the environment.
\begin{table}[t]
    \centering
    \caption{Task Performance on User Study.}
    \label{tab:ux-performance}
    \vspace{-10pt}
    \begin{threeparttable}
        \begin{tabular}{ccccc}
            \toprule
            & $r_\mathrm{tidy}$ & $r_\mathrm{serve}$ & $t_\mathrm{tidy}$ & $t_\mathrm{serve}$ \\
            \midrule
            Ours & 13/14 & 6/13 & 923 & 830\\
            Baseline & 13/14 & 11/13 & 802 & 463 \\
            \bottomrule
        \end{tabular}
    \end{threeparttable}
    \vspace{-10pt}
\end{table}

\section{CONCLUSIONS}
We proposed Sketch-MoMa, a teleoperation system for mobile manipulation via hand-drawn sketches.
We validated the capability of VLMs with 7 tasks and 5 sketch shapes.
Our method was effective in specifying motion details using shape and word sketches as visual clues.
Our user study showed 
that sketches have the potential to be an easy and intuitive interface.
Our future goal is to apply various tasks based on users' varied sketches and stabilize the system to increase usability, enabling users to operate the robot anytime and anywhere as intended.
\section*{ACKNOWLEDGMENT}
We thank Masahiro Kagi, Taro Kono, and Kota Shinjo for their contributions to the creation of the evaluation tool and Shigemichi Matsuzaki for his contribution to proofreading and reviewing of this paper. 
This research was partially supported by grants from the Ichihara International Scholarship Foundation and The Nitto Foundation.

\printbibliography

@inproceedings{dipalo2024keypointactiontokens,
   author = {Norman Palo and Edward Johns},
   doi = {10.15607/RSS.2024.XX.096},
   isbn = {979-8-9902848-0-7},
   booktitle = {Robotics: Science and Systems XX},
   month = {7},
   publisher = {Robotics: Science and Systems Foundation},
   title = {Keypoint Action Tokens Enable In-Context Imitation Learning in Robotics},
   year = {2024},
}

@inproceedings{liang2023codepolicieslanguagemodel,
   author = {Jacky Liang and Wenlong Huang and Fei Xia and Peng Xu and Karol Hausman and Brian Ichter and Pete Florence and Andy Zeng},
   doi = {10.1109/ICRA48891.2023.10160591},
   booktitle = {2023 IEEE International Conference on Robotics and Automation (ICRA)},
   month = {5},
   pages = {9493-9500},
   publisher = {IEEE},
   title = {Code as Policies: Language Model Programs for Embodied Control},
   url = {https://ieeexplore.ieee.org/document/10160591/},
   year = {2023},
}

@inproceedings{nasiriany2024pivotiterativevisualprompting,
   author = {Soroush Nasiriany and Fei Xia and Wenhao Yu and Ted Xiao and Jacky Liang and Ishita Dasgupta and Annie Xie and Danny Driess and Ayzaan Wahid and Zhuo Xu and Quan Vuong and Tingnan Zhang and Tsang-Wei Edward Lee and Kuang-Huei Lee and Peng Xu and Sean Kirmani and Yuke Zhu and Andy Zeng and Karol Hausman and Nicolas Heess and Chelsea Finn and Sergey Levine and brian ichter},
   booktitle = {Forty-first International Conference on Machine Learning},
   title = {PIVOT: Iterative Visual Prompting Elicits Actionable Knowledge for VLMs},
   url = {https://openreview.net/forum?id=051jaf8MQy},
   year = {2024},
}

@inproceedings{rana2023sayplangroundinglargelanguage,
   author = {Krishan Rana and Jesse Haviland and Sourav Garg and Jad Abou-Chakra and Ian Reid and Niko Suenderhauf},
   booktitle = {7th Annual Conference on Robot Learning},
   title = {SayPlan: Grounding Large Language Models using 3D Scene Graphs for Scalable Robot Task Planning},
   url = {https://openreview.net/forum?id=wMpOMO0Ss7a},
   year = {2023},
}

@inproceedings{huang2023voxposercomposable3dvalue,
   author = {Wenlong Huang and Chen Wang and Ruohan Zhang and Yunzhu Li and Jiajun Wu and Li Fei-Fei},
   booktitle = {7th Annual Conference on Robot Learning},
   month = {7},
   title = {VoxPoser: Composable 3D Value Maps for Robotic Manipulation with Language Models},
   url = {http://arxiv.org/abs/2307.05973},
   year = {2023},
}

@inproceedings{rt22023arxiv,
   author = {Brianna Zitkovich and Tianhe Yu and Sichun Xu and Peng Xu and Ted Xiao and Fei Xia and Jialin Wu and Paul Wohlhart and Stefan Welker and Ayzaan Wahid and Quan Vuong and Vincent Vanhoucke and Huong Tran and Radu Soricut and Anikait Singh and Jaspiar Singh and Pierre Sermanet and Pannag R Sanketi and Grecia Salazar and Michael S Ryoo and Krista Reymann and Kanishka Rao and Karl Pertsch and Igor Mordatch and Henryk Michalewski and Yao Lu and Sergey Levine and Lisa Lee and Tsang-Wei Edward Lee and Isabel Leal and Yuheng Kuang and Dmitry Kalashnikov and Ryan Julian and Nikhil J Joshi and Alex Irpan and brian ichter and Jasmine Hsu and Alexander Herzog and Karol Hausman and Keerthana Gopalakrishnan and Chuyuan Fu and Pete Florence and Chelsea Finn and Kumar Avinava Dubey and Danny Driess and Tianli Ding and Krzysztof Marcin Choromanski and Xi Chen and Yevgen Chebotar and Justice Carbajal and Noah Brown and Anthony Brohan and Montserrat Gonzalez Arenas and Kehang Han},
   booktitle = {7th Annual Conference on Robot Learning},
   title = {RT-2: Vision-Language-Action Models Transfer Web Knowledge to Robotic Control},
   url = {https://openreview.net/forum?id=XMQgwiJ7KSX},
   year = {2023},
}

@inproceedings{liu2024mokaopenvocabularyroboticmanipulation,
   author = {Kuan Fang and Fangchen Liu and Pieter Abbeel and Sergey Levine},
   doi = {10.15607/RSS.2024.XX.062},
   isbn = {979-8-9902848-0-7},
   booktitle = {Robotics: Science and Systems XX},
   month = {7},
   publisher = {Robotics: Science and Systems Foundation},
   title = {MOKA: Open-World Robotic Manipulation through Mark-Based Visual Prompting},
   url = {http://www.roboticsproceedings.org/rss20/p062.pdf},
   year = {2024},
}

@inproceedings{brohan2023rt1roboticstransformerrealworld,
   author = {Anthony Brohan and Noah Brown and Justice Carbajal and Yevgen Chebotar and Joseph Dabis and Chelsea Finn and Keerthana Gopalakrishnan and Karol Hausman and Alexander Herzog and Jasmine Hsu and Julian Ibarz and Brian Ichter and Alex Irpan and Tomas Jackson and Sally Jesmonth and Nikhil Joshi and Ryan Julian and Dmitry Kalashnikov and Yuheng Kuang and Isabel Leal and Kuang-Huei Lee and Sergey Levine and Yao Lu and Utsav Malla and Deeksha Manjunath and Igor Mordatch and Ofir Nachum and Carolina Parada and Jodilyn Peralta and Emily Perez and Karl Pertsch and Jornell Quiambao and Kanishka Rao and Michael Ryoo and Grecia Salazar and Pannag Sanketi and Kevin Sayed and Jaspiar Singh and Sumedh Sontakke and Austin Stone and Clayton Tan and Huong Tran and Vincent Vanhoucke and Steve Vega and Quan Vuong and Fei Xia and Ted Xiao and Peng Xu and Sichun Xu and Tianhe Yu and Brianna Zitkovich},
   doi = {10.15607/RSS.2023.XIX.025},
   isbn = {978-0-9923747-9-2},
   booktitle = {Robotics: Science and Systems XIX},
   month = {7},
   publisher = {Robotics: Science and Systems Foundation},
   title = {RT-1: Robotics Transformer for Real-World Control at Scale},
   url = {http://www.roboticsproceedings.org/rss19/p025.pdf},
   year = {2023},
}

@article{firoozi2023foundationmodelsroboticsapplications,
   author = {Roya Firoozi and Johnathan Tucker and Stephen Tian and Anirudha Majumdar and Jiankai Sun and Weiyu Liu and Yuke Zhu and Shuran Song and Ashish Kapoor and Karol Hausman and Brian Ichter and Danny Driess and Jiajun Wu and Cewu Lu and Mac Schwager},
   doi = {10.1177/02783649241281508},
   issn = {0278-3649},
   journal = {The International Journal of Robotics Research},
   month = {9},
   title = {Foundation models in robotics: Applications, challenges, and the future},
   url = {https://journals.sagepub.com/doi/10.1177/02783649241281508},
   year = {2024},
}

@techreport{hu2023generalpurposerobotsfoundationmodels,
   author = {Yafei Hu and Quanting Xie and Vidhi Jain and Jonathan Francis and Jay Patrikar and Nikhil Keetha and Seungchan Kim and Yaqi Xie and Tianyi Zhang and Hao-Shu Fang and Shibo Zhao and Shayegan Omidshafiei and Dong-Ki Kim and Ali-akbar Agha-mohammadi and Katia Sycara and Matthew Johnson-Roberson and Dhruv Batra and Xiaolong Wang and Sebastian Scherer and Chen Wang and Zsolt Kira and Fei Xia and Yonatan Bisk},
   month = {12},
   title = {Toward General-Purpose Robots via Foundation Models: A Survey and Meta-Analysis},
   url = {http://arxiv.org/abs/2312.08782},
   year = {2023},
}

@article{kawaharazuka2024realworldrobotapplicationsfoundation,
author = {Kento Kawaharazuka and Tatsuya Matsushima and Andrew Gambardella and Jiaxian Guo and Chris Paxton and Andy Zeng},
title = {Real-world robot applications of foundation models: a review},
journal = {Advanced Robotics},
volume = {38},
number = {18},
pages = {1232--1254},
year = {2024},
publisher = {Taylor \& Francis},
doi = {10.1080/01691864.2024.2408593},


URL = { 
    
        https://doi.org/10.1080/01691864.2024.2408593
    
    

},
eprint = { 
    
        https://doi.org/10.1080/01691864.2024.2408593
    
    

}

}

@inproceedings{embodimentcollaboration2024openxembodimentroboticlearning,
   author = {Abby O’Neill and Abdul Rehman and Abhiram Maddukuri and Abhishek Gupta and Abhishek Padalkar and Abraham Lee and Acorn Pooley and Agrim Gupta and Ajay Mandlekar and Ajinkya Jain and Albert Tung and Alex Bewley and Alex Herzog and Alex Irpan and Alexander Khazatsky and Anant Rai and Anchit Gupta and Andrew Wang and Anikait Singh and Animesh Garg and Aniruddha Kembhavi and Annie Xie and Anthony Brohan and Antonin Raffin and Archit Sharma and Arefeh Yavary and Arhan Jain and Ashwin Balakrishna and Ayzaan Wahid and Ben Burgess-Limerick and Beomjoon Kim and Bernhard Schölkopf and Blake Wulfe and Brian Ichter and Cewu Lu and Charles Xu and Charlotte Le and Chelsea Finn and Chen Wang and Chenfeng Xu and Cheng Chi and Chenguang Huang and Christine Chan and Christopher Agia and Chuer Pan and Chuyuan Fu and Coline Devin and Danfei Xu and Daniel Morton and Danny Driess and Daphne Chen and Deepak Pathak and Dhruv Shah and Dieter Büchler and Dinesh Jayaraman and Dmitry Kalashnikov and Dorsa Sadigh and Edward Johns and Ethan Foster and Fangchen Liu and Federico Ceola and Fei Xia and Feiyu Zhao and Freek Stulp and Gaoyue Zhou and Gaurav S. Sukhatme and Gautam Salhotra and Ge Yan and Gilbert Feng and Giulio Schiavi and Glen Berseth and Gregory Kahn and Guanzhi Wang and Hao Su and Hao-Shu Fang and Haochen Shi and Henghui Bao and Heni Ben Amor and Henrik I Christensen and Hiroki Furuta and Homer Walke and Hongjie Fang and Huy Ha and Igor Mordatch and Ilija Radosavovic and Isabel Leal and Jacky Liang and Jad Abou-Chakra and Jaehyung Kim and Jaimyn Drake and Jan Peters and Jan Schneider and Jasmine Hsu and Jeannette Bohg and Jeffrey Bingham and Jeffrey Wu and Jensen Gao and Jiaheng Hu and Jiajun Wu and Jialin Wu and Jiankai Sun and Jianlan Luo and Jiayuan Gu and Jie Tan and Jihoon Oh and Jimmy Wu and Jingpei Lu and Jingyun Yang and Jitendra Malik and João Silvério and Joey Hejna and Jonathan Booher and Jonathan Tompson and Jonathan Yang and Jordi Salvador and Joseph J. Lim and Junhyek Han and Kaiyuan Wang and Kanishka Rao and Karl Pertsch and Karol Hausman and Keegan Go and Keerthana Gopalakrishnan and Ken Goldberg and Kendra Byrne and Kenneth Oslund and Kento Kawaharazuka and Kevin Black and Kevin Lin and Kevin Zhang and Kiana Ehsani and Kiran Lekkala and Kirsty Ellis and Krishan Rana and Krishnan Srinivasan and Kuan Fang and Kunal Pratap Singh and Kuo-Hao Zeng and Kyle Hatch and Kyle Hsu and Laurent Itti and Lawrence Yunliang Chen and Lerrel Pinto and Li Fei-Fei and Liam Tan and Linxi Jim Fan and Lionel Ott and Lisa Lee and Luca Weihs and Magnum Chen and Marion Lepert and Marius Memmel and Masayoshi Tomizuka and Masha Itkina and Mateo Guaman Castro and Max Spero and Maximilian Du and Michael Ahn and Michael C. Yip and Mingtong Zhang and Mingyu Ding and Minho Heo and Mohan Kumar Srirama and Mohit Sharma and Moo Jin Kim and Naoaki Kanazawa and Nicklas Hansen and Nicolas Heess and Nikhil J Joshi and Niko Suenderhauf and Ning Liu and Norman Di Palo and Nur Muhammad Mahi Shafiullah and Oier Mees and Oliver Kroemer and Osbert Bastani and Pannag R Sanketi and Patrick Tree Miller and Patrick Yin and Paul Wohlhart and Peng Xu and Peter David Fagan and Peter Mitrano and Pierre Sermanet and Pieter Abbeel and Priya Sundaresan and Qiuyu Chen and Quan Vuong and Rafael Rafailov and Ran Tian and Ria Doshi and Roberto Martín-Martín and Rohan Baijal and Rosario Scalise and Rose Hendrix and Roy Lin and Runjia Qian and Ruohan Zhang and Russell Mendonca and Rutav Shah and Ryan Hoque and Ryan Julian and Samuel Bustamante and Sean Kirmani and Sergey Levine and Shan Lin and Sherry Moore and Shikhar Bahl and Shivin Dass and Shubham Sonawani and Shuran Song and Sichun Xu and Siddhant Haldar and Siddharth Karamcheti and Simeon Adebola and Simon Guist and Soroush Nasiriany and Stefan Schaal and Stefan Welker and Stephen Tian and Subramanian Ramamoorthy and Sudeep Dasari and Suneel Belkhale and Sungjae Park and Suraj Nair and Suvir Mirchandani and Takayuki Osa and Tanmay Gupta and Tatsuya Harada and Tatsuya Matsushima and Ted Xiao and Thomas Kollar and Tianhe Yu and Tianli Ding and Todor Davchev and Tony Z. Zhao and Travis Armstrong and Trevor Darrell and Trinity Chung and Vidhi Jain and Vincent Vanhoucke and Wei Zhan and Wenxuan Zhou and Wolfram Burgard and Xi Chen and Xiaolong Wang and Xinghao Zhu and Xinyang Geng and Xiyuan Liu and Xu Liangwei and Xuanlin Li and Yao Lu and Yecheng Jason Ma and Yejin Kim and Yevgen Chebotar and Yifan Zhou and Yifeng Zhu and Yilin Wu and Ying Xu and Yixuan Wang and Yonatan Bisk and Yoonyoung Cho and Youngwoon Lee and Yuchen Cui and Yue Cao and Yueh-Hua Wu and Yujin Tang and Yuke Zhu and Yunchu Zhang and Yunfan Jiang and Yunshuang Li and Yunzhu Li and Yusuke Iwasawa and Yutaka Matsuo and Zehan Ma and Zhuo Xu and Zichen Jeff Cui and Zichen Zhang and Zipeng Lin},
   doi = {10.1109/ICRA57147.2024.10611477},
   isbn = {979-8-3503-8457-4},
   booktitle = {2024 IEEE International Conference on Robotics and Automation (ICRA)},
   month = {5},
   pages = {6892-6903},
   publisher = {IEEE},
   title = {Open X-Embodiment: Robotic Learning Datasets and RT-X Models : Open X-Embodiment Collaboration},
   url = {https://ieeexplore.ieee.org/document/10611477/},
   year = {2024},
}

@techreport{gpt4,
   author = {OpenAI and Josh Achiam and Steven Adler and Sandhini Agarwal and Lama Ahmad and Ilge Akkaya and Florencia Leoni Aleman and Diogo Almeida and Janko Altenschmidt and Sam Altman and Shyamal Anadkat and Red Avila and Igor Babuschkin and Suchir Balaji and Valerie Balcom and Paul Baltescu and Haiming Bao and Mohammad Bavarian and Jeff Belgum and Irwan Bello and Jake Berdine and Gabriel Bernadett-Shapiro and Christopher Berner and Lenny Bogdonoff and Oleg Boiko and Madelaine Boyd and Anna-Luisa Brakman and Greg Brockman and Tim Brooks and Miles Brundage and Kevin Button and Trevor Cai and Rosie Campbell and Andrew Cann and Brittany Carey and Chelsea Carlson and Rory Carmichael and Brooke Chan and Che Chang and Fotis Chantzis and Derek Chen and Sully Chen and Ruby Chen and Jason Chen and Mark Chen and Ben Chess and Chester Cho and Casey Chu and Hyung Won Chung and Dave Cummings and Jeremiah Currier and Yunxing Dai and Cory Decareaux and Thomas Degry and Noah Deutsch and Damien Deville and Arka Dhar and David Dohan and Steve Dowling and Sheila Dunning and Adrien Ecoffet and Atty Eleti and Tyna Eloundou and David Farhi and Liam Fedus and Niko Felix and Simón Posada Fishman and Juston Forte and Isabella Fulford and Leo Gao and Elie Georges and Christian Gibson and Vik Goel and Tarun Gogineni and Gabriel Goh and Rapha Gontijo-Lopes and Jonathan Gordon and Morgan Grafstein and Scott Gray and Ryan Greene and Joshua Gross and Shixiang Shane Gu and Yufei Guo and Chris Hallacy and Jesse Han and Jeff Harris and Yuchen He and Mike Heaton and Johannes Heidecke and Chris Hesse and Alan Hickey and Wade Hickey and Peter Hoeschele and Brandon Houghton and Kenny Hsu and Shengli Hu and Xin Hu and Joost Huizinga and Shantanu Jain and Shawn Jain and Joanne Jang and Angela Jiang and Roger Jiang and Haozhun Jin and Denny Jin and Shino Jomoto and Billie Jonn and Heewoo Jun and Tomer Kaftan and Łukasz Kaiser and Ali Kamali and Ingmar Kanitscheider and Nitish Shirish Keskar and Tabarak Khan and Logan Kilpatrick and Jong Wook Kim and Christina Kim and Yongjik Kim and Jan Hendrik Kirchner and Jamie Kiros and Matt Knight and Daniel Kokotajlo and Łukasz Kondraciuk and Andrew Kondrich and Aris Konstantinidis and Kyle Kosic and Gretchen Krueger and Vishal Kuo and Michael Lampe and Ikai Lan and Teddy Lee and Jan Leike and Jade Leung and Daniel Levy and Chak Ming Li and Rachel Lim and Molly Lin and Stephanie Lin and Mateusz Litwin and Theresa Lopez and Ryan Lowe and Patricia Lue and Anna Makanju and Kim Malfacini and Sam Manning and Todor Markov and Yaniv Markovski and Bianca Martin and Katie Mayer and Andrew Mayne and Bob McGrew and Scott Mayer McKinney and Christine McLeavey and Paul McMillan and Jake McNeil and David Medina and Aalok Mehta and Jacob Menick and Luke Metz and Andrey Mishchenko and Pamela Mishkin and Vinnie Monaco and Evan Morikawa and Daniel Mossing and Tong Mu and Mira Murati and Oleg Murk and David Mély and Ashvin Nair and Reiichiro Nakano and Rajeev Nayak and Arvind Neelakantan and Richard Ngo and Hyeonwoo Noh and Long Ouyang and Cullen O'Keefe and Jakub Pachocki and Alex Paino and Joe Palermo and Ashley Pantuliano and Giambattista Parascandolo and Joel Parish and Emy Parparita and Alex Passos and Mikhail Pavlov and Andrew Peng and Adam Perelman and Filipe de Avila Belbute Peres and Michael Petrov and Henrique Ponde de Oliveira Pinto and Michael and Pokorny and Michelle Pokrass and Vitchyr H. Pong and Tolly Powell and Alethea Power and Boris Power and Elizabeth Proehl and Raul Puri and Alec Radford and Jack Rae and Aditya Ramesh and Cameron Raymond and Francis Real and Kendra Rimbach and Carl Ross and Bob Rotsted and Henri Roussez and Nick Ryder and Mario Saltarelli and Ted Sanders and Shibani Santurkar and Girish Sastry and Heather Schmidt and David Schnurr and John Schulman and Daniel Selsam and Kyla Sheppard and Toki Sherbakov and Jessica Shieh and Sarah Shoker and Pranav Shyam and Szymon Sidor and Eric Sigler and Maddie Simens and Jordan Sitkin and Katarina Slama and Ian Sohl and Benjamin Sokolowsky and Yang Song and Natalie Staudacher and Felipe Petroski Such and Natalie Summers and Ilya Sutskever and Jie Tang and Nikolas Tezak and Madeleine B. Thompson and Phil Tillet and Amin Tootoonchian and Elizabeth Tseng and Preston Tuggle and Nick Turley and Jerry Tworek and Juan Felipe Cerón Uribe and Andrea Vallone and Arun Vijayvergiya and Chelsea Voss and Carroll Wainwright and Justin Jay Wang and Alvin Wang and Ben Wang and Jonathan Ward and Jason Wei and CJ Weinmann and Akila Welihinda and Peter Welinder and Jiayi Weng and Lilian Weng and Matt Wiethoff and Dave Willner and Clemens Winter and Samuel Wolrich and Hannah Wong and Lauren Workman and Sherwin Wu and Jeff Wu and Michael Wu and Kai Xiao and Tao Xu and Sarah Yoo and Kevin Yu and Qiming Yuan and Wojciech Zaremba and Rowan Zellers and Chong Zhang and Marvin Zhang and Shengjia Zhao and Tianhao Zheng and Juntang Zhuang and William Zhuk and Barret Zoph},
   month = {3},
   title = {GPT-4 Technical Report},
   url = {http://arxiv.org/abs/2303.08774},
   year = {2023},
}

@techreport{gemini,
   author = {GeminiTeam and Rohan Anil and Sebastian Borgeaud and Jean-Baptiste Alayrac and Jiahui Yu and Radu Soricut and Johan Schalkwyk and Andrew M. Dai and Anja Hauth and Katie Millican and David Silver and Melvin Johnson and Ioannis Antonoglou and Julian Schrittwieser and Amelia Glaese and Jilin Chen and Emily Pitler and Timothy Lillicrap and Angeliki Lazaridou and Orhan Firat and James Molloy and Michael Isard and Paul R. Barham and Tom Hennigan and Benjamin Lee and Fabio Viola and Malcolm Reynolds and Yuanzhong Xu and Ryan Doherty and Eli Collins and Clemens Meyer and Eliza Rutherford and Erica Moreira and Kareem Ayoub and Megha Goel and Jack Krawczyk and Cosmo Du and Ed Chi and Heng-Tze Cheng and Eric Ni and Purvi Shah and Patrick Kane and Betty Chan and Manaal Faruqui and Aliaksei Severyn and Hanzhao Lin and YaGuang Li and Yong Cheng and Abe Ittycheriah and Mahdis Mahdieh and Mia Chen and Pei Sun and Dustin Tran and Sumit Bagri and Balaji Lakshminarayanan and Jeremiah Liu and Andras Orban and Fabian Güra and Hao Zhou and Xinying Song and Aurelien Boffy and Harish Ganapathy and Steven Zheng and HyunJeong Choe and Ágoston Weisz and Tao Zhu and Yifeng Lu and Siddharth Gopal and Jarrod Kahn and Maciej Kula and Jeff Pitman and Rushin Shah and Emanuel Taropa and Majd Al Merey and Martin Baeuml and Zhifeng Chen and Laurent El Shafey and Yujing Zhang and Olcan Sercinoglu and George Tucker and Enrique Piqueras and Maxim Krikun and Iain Barr and Nikolay Savinov and Ivo Danihelka and Becca Roelofs and Anaïs White and Anders Andreassen and Tamara von Glehn and Lakshman Yagati and Mehran Kazemi and Lucas Gonzalez and Misha Khalman and Jakub Sygnowski and Alexandre Frechette and Charlotte Smith and Laura Culp and Lev Proleev and Yi Luan and Xi Chen and James Lottes and Nathan Schucher and Federico Lebron and Alban Rrustemi and Natalie Clay and Phil Crone and Tomas Kocisky and Jeffrey Zhao and Bartek Perz and Dian Yu and Heidi Howard and Adam Bloniarz and Jack W. Rae and Han Lu and Laurent Sifre and Marcello Maggioni and Fred Alcober and Dan Garrette and Megan Barnes and Shantanu Thakoor and Jacob Austin and Gabriel Barth-Maron and William Wong and Rishabh Joshi and Rahma Chaabouni and Deeni Fatiha and Arun Ahuja and Gaurav Singh Tomar and Evan Senter and Martin Chadwick and Ilya Kornakov and Nithya Attaluri and Iñaki Iturrate and Ruibo Liu and Yunxuan Li and Sarah Cogan and Jeremy Chen and Chao Jia and Chenjie Gu and Qiao Zhang and Jordan Grimstad and Ale Jakse Hartman and Xavier Garcia and Thanumalayan Sankaranarayana Pillai and Jacob Devlin and Michael Laskin and Diego de Las Casas and Dasha Valter and Connie Tao and Lorenzo Blanco and Adrià Puigdomènech Badia and David Reitter and Mianna Chen and Jenny Brennan and Clara Rivera and Sergey Brin and Shariq Iqbal and Gabriela Surita and Jane Labanowski and Abhi Rao and Stephanie Winkler and Emilio Parisotto and Yiming Gu and Kate Olszewska and Ravi Addanki and Antoine Miech and Annie Louis and Denis Teplyashin and Geoff Brown and Elliot Catt and Jan Balaguer and Jackie Xiang and Pidong Wang and Zoe Ashwood and Anton Briukhov and Albert Webson and Sanjay Ganapathy and Smit Sanghavi and Ajay Kannan and Ming-Wei Chang and Axel Stjerngren and Josip Djolonga and Yuting Sun and Ankur Bapna and Matthew Aitchison and Pedram Pejman and Henryk Michalewski and Tianhe Yu and Cindy Wang and Juliette Love and Junwhan Ahn and Dawn Bloxwich and Kehang Han and Peter Humphreys and Thibault Sellam and James Bradbury and Varun Godbole and Sina Samangooei and Bogdan Damoc and Alex Kaskasoli and Sébastien M. R. Arnold and Vijay Vasudevan and Shubham Agrawal and Jason Riesa and Dmitry Lepikhin and Richard Tanburn and Srivatsan Srinivasan and Hyeontaek Lim and Sarah Hodkinson and Pranav Shyam and Johan Ferret and Steven Hand and Ankush Garg and Tom Le Paine and Jian Li and Yujia Li and Minh Giang and Alexander Neitz and Zaheer Abbas and Sarah York and Machel Reid and Elizabeth Cole and Aakanksha Chowdhery and Dipanjan Das and Dominika Rogozińska and Vitaliy Nikolaev and Pablo Sprechmann and Zachary Nado and Lukas Zilka and Flavien Prost and Luheng He and Marianne Monteiro and Gaurav Mishra and Chris Welty and Josh Newlan and Dawei Jia and Miltiadis Allamanis and Clara Huiyi Hu and Raoul de Liedekerke and Justin Gilmer and Carl Saroufim and Shruti Rijhwani and Shaobo Hou and Disha Shrivastava and Anirudh Baddepudi and Alex Goldin and Adnan Ozturel and Albin Cassirer and Yunhan Xu and Daniel Sohn and Devendra Sachan and Reinald Kim Amplayo and Craig Swanson and Dessie Petrova and Shashi Narayan and Arthur Guez and Siddhartha Brahma and Jessica Landon and Miteyan Patel and Ruizhe Zhao and Kevin Villela and Luyu Wang and Wenhao Jia and Matthew Rahtz and Mai Giménez and Legg Yeung and James Keeling and Petko Georgiev and Diana Mincu and Boxi Wu and Salem Haykal and Rachel Saputro and Kiran Vodrahalli and James Qin and Zeynep Cankara and Abhanshu Sharma and Nick Fernando and Will Hawkins and Behnam Neyshabur and Solomon Kim and Adrian Hutter and Priyanka Agrawal and Alex Castro-Ros and George van den Driessche and Tao Wang and Fan Yang and Shuo-yiin Chang and Paul Komarek and Ross McIlroy and Mario Lučić and Guodong Zhang and Wael Farhan and Michael Sharman and Paul Natsev and Paul Michel and Yamini Bansal and Siyuan Qiao and Kris Cao and Siamak Shakeri and Christina Butterfield and Justin Chung and Paul Kishan Rubenstein and Shivani Agrawal and Arthur Mensch and Kedar Soparkar and Karel Lenc and Timothy Chung and Aedan Pope and Loren Maggiore and Jackie Kay and Priya Jhakra and Shibo Wang and Joshua Maynez and Mary Phuong and Taylor Tobin and Andrea Tacchetti and Maja Trebacz and Kevin Robinson and Yash Katariya and Sebastian Riedel and Paige Bailey and Kefan Xiao and Nimesh Ghelani and Lora Aroyo and Ambrose Slone and Neil Houlsby and Xuehan Xiong and Zhen Yang and Elena Gribovskaya and Jonas Adler and Mateo Wirth and Lisa Lee and Music Li and Thais Kagohara and Jay Pavagadhi and Sophie Bridgers and Anna Bortsova and Sanjay Ghemawat and Zafarali Ahmed and Tianqi Liu and Richard Powell and Vijay Bolina and Mariko Iinuma and Polina Zablotskaia and James Besley and Da-Woon Chung and Timothy Dozat and Ramona Comanescu and Xiance Si and Jeremy Greer and Guolong Su and Martin Polacek and Raphaël Lopez Kaufman and Simon Tokumine and Hexiang Hu and Elena Buchatskaya and Yingjie Miao and Mohamed Elhawaty and Aditya Siddhant and Nenad Tomasev and Jinwei Xing and Christina Greer and Helen Miller and Shereen Ashraf and Aurko Roy and Zizhao Zhang and Ada Ma and Angelos Filos and Milos Besta and Rory Blevins and Ted Klimenko and Chih-Kuan Yeh and Soravit Changpinyo and Jiaqi Mu and Oscar Chang and Mantas Pajarskas and Carrie Muir and Vered Cohen and Charline Le Lan and Krishna Haridasan and Amit Marathe and Steven Hansen and Sholto Douglas and Rajkumar Samuel and Mingqiu Wang and Sophia Austin and Chang Lan and Jiepu Jiang and Justin Chiu and Jaime Alonso Lorenzo and Lars Lowe Sjösund and Sébastien Cevey and Zach Gleicher and Thi Avrahami and Anudhyan Boral and Hansa Srinivasan and Vittorio Selo and Rhys May and Konstantinos Aisopos and Léonard Hussenot and Livio Baldini Soares and Kate Baumli and Michael B. Chang and Adrià Recasens and Ben Caine and Alexander Pritzel and Filip Pavetic and Fabio Pardo and Anita Gergely and Justin Frye and Vinay Ramasesh and Dan Horgan and Kartikeya Badola and Nora Kassner and Subhrajit Roy and Ethan Dyer and Víctor Campos Campos and Alex Tomala and Yunhao Tang and Dalia El Badawy and Elspeth White and Basil Mustafa and Oran Lang and Abhishek Jindal and Sharad Vikram and Zhitao Gong and Sergi Caelles and Ross Hemsley and Gregory Thornton and Fangxiaoyu Feng and Wojciech Stokowiec and Ce Zheng and Phoebe Thacker and Çağlar Ünlü and Zhishuai Zhang and Mohammad Saleh and James Svensson and Max Bileschi and Piyush Patil and Ankesh Anand and Roman Ring and Katerina Tsihlas and Arpi Vezer and Marco Selvi and Toby Shevlane and Mikel Rodriguez and Tom Kwiatkowski and Samira Daruki and Keran Rong and Allan Dafoe and Nicholas FitzGerald and Keren Gu-Lemberg and Mina Khan and Lisa Anne Hendricks and Marie Pellat and Vladimir Feinberg and James Cobon-Kerr and Tara Sainath and Maribeth Rauh and Sayed Hadi Hashemi and Richard Ives and Yana Hasson and Eric Noland and Yuan Cao and Nathan Byrd and Le Hou and Qingze Wang and Thibault Sottiaux and Michela Paganini and Jean-Baptiste Lespiau and Alexandre Moufarek and Samer Hassan and Kaushik Shivakumar and Joost van Amersfoort and Amol Mandhane and Pratik Joshi and Anirudh Goyal and Matthew Tung and Andrew Brock and Hannah Sheahan and Vedant Misra and Cheng Li and Nemanja Rakićević and Mostafa Dehghani and Fangyu Liu and Sid Mittal and Junhyuk Oh and Seb Noury and Eren Sezener and Fantine Huot and Matthew Lamm and Nicola De Cao and Charlie Chen and Sidharth Mudgal and Romina Stella and Kevin Brooks and Gautam Vasudevan and Chenxi Liu and Mainak Chain and Nivedita Melinkeri and Aaron Cohen and Venus Wang and Kristie Seymore and Sergey Zubkov and Rahul Goel and Summer Yue and Sai Krishnakumaran and Brian Albert and Nate Hurley and Motoki Sano and Anhad Mohananey and Jonah Joughin and Egor Filonov and Tomasz Kępa and Yomna Eldawy and Jiawern Lim and Rahul Rishi and Shirin Badiezadegan and Taylor Bos and Jerry Chang and Sanil Jain and Sri Gayatri Sundara Padmanabhan and Subha Puttagunta and Kalpesh Krishna and Leslie Baker and Norbert Kalb and Vamsi Bedapudi and Adam Kurzrok and Shuntong Lei and Anthony Yu and Oren Litvin and Xiang Zhou and Zhichun Wu and Sam Sobell and Andrea Siciliano and Alan Papir and Robby Neale and Jonas Bragagnolo and Tej Toor and Tina Chen and Valentin Anklin and Feiran Wang and Richie Feng and Milad Gholami and Kevin Ling and Lijuan Liu and Jules Walter and Hamid Moghaddam and Arun Kishore and Jakub Adamek and Tyler Mercado and Jonathan Mallinson and Siddhinita Wandekar and Stephen Cagle and Eran Ofek and Guillermo Garrido and Clemens Lombriser and Maksim Mukha and Botu Sun and Hafeezul Rahman Mohammad and Josip Matak and Yadi Qian and Vikas Peswani and Pawel Janus and Quan Yuan and Leif Schelin and Oana David and Ankur Garg and Yifan He and Oleksii Duzhyi and Anton Älgmyr and Timothée Lottaz and Qi Li and Vikas Yadav and Luyao Xu and Alex Chinien and Rakesh Shivanna and Aleksandr Chuklin and Josie Li and Carrie Spadine and Travis Wolfe and Kareem Mohamed and Subhabrata Das and Zihang Dai and Kyle He and Daniel von Dincklage and Shyam Upadhyay and Akanksha Maurya and Luyan Chi and Sebastian Krause and Khalid Salama and Pam G Rabinovitch and Pavan Kumar Reddy M and Aarush Selvan and Mikhail Dektiarev and Golnaz Ghiasi and Erdem Guven and Himanshu Gupta and Boyi Liu and Deepak Sharma and Idan Heimlich Shtacher and Shachi Paul and Oscar Akerlund and François-Xavier Aubet and Terry Huang and Chen Zhu and Eric Zhu and Elico Teixeira and Matthew Fritze and Francesco Bertolini and Liana-Eleonora Marinescu and Martin Bölle and Dominik Paulus and Khyatti Gupta and Tejasi Latkar and Max Chang and Jason Sanders and Roopa Wilson and Xuewei Wu and Yi-Xuan Tan and Lam Nguyen Thiet and Tulsee Doshi and Sid Lall and Swaroop Mishra and Wanming Chen and Thang Luong and Seth Benjamin and Jasmine Lee and Ewa Andrejczuk and Dominik Rabiej and Vipul Ranjan and Krzysztof Styrc and Pengcheng Yin and Jon Simon and Malcolm Rose Harriott and Mudit Bansal and Alexei Robsky and Geoff Bacon and David Greene and Daniil Mirylenka and Chen Zhou and Obaid Sarvana and Abhimanyu Goyal and Samuel Andermatt and Patrick Siegler and Ben Horn and Assaf Israel and Francesco Pongetti and Chih-Wei "Louis" Chen and Marco Selvatici and Pedro Silva and Kathie Wang and Jackson Tolins and Kelvin Guu and Roey Yogev and Xiaochen Cai and Alessandro Agostini and Maulik Shah and Hung Nguyen and Noah Ó Donnaile and Sébastien Pereira and Linda Friso and Adam Stambler and Adam Kurzrok and Chenkai Kuang and Yan Romanikhin and Mark Geller and ZJ Yan and Kane Jang and Cheng-Chun Lee and Wojciech Fica and Eric Malmi and Qijun Tan and Dan Banica and Daniel Balle and Ryan Pham and Yanping Huang and Diana Avram and Hongzhi Shi and Jasjot Singh and Chris Hidey and Niharika Ahuja and Pranab Saxena and Dan Dooley and Srividya Pranavi Potharaju and Eileen O'Neill and Anand Gokulchandran and Ryan Foley and Kai Zhao and Mike Dusenberry and Yuan Liu and Pulkit Mehta and Ragha Kotikalapudi and Chalence Safranek-Shrader and Andrew Goodman and Joshua Kessinger and Eran Globen and Prateek Kolhar and Chris Gorgolewski and Ali Ibrahim and Yang Song and Ali Eichenbaum and Thomas Brovelli and Sahitya Potluri and Preethi Lahoti and Cip Baetu and Ali Ghorbani and Charles Chen and Andy Crawford and Shalini Pal and Mukund Sridhar and Petru Gurita and Asier Mujika and Igor Petrovski and Pierre-Louis Cedoz and Chenmei Li and Shiyuan Chen and Niccolò Dal Santo and Siddharth Goyal and Jitesh Punjabi and Karthik Kappaganthu and Chester Kwak and Pallavi LV and Sarmishta Velury and Himadri Choudhury and Jamie Hall and Premal Shah and Ricardo Figueira and Matt Thomas and Minjie Lu and Ting Zhou and Chintu Kumar and Thomas Jurdi and Sharat Chikkerur and Yenai Ma and Adams Yu and Soo Kwak and Victor Ähdel and Sujeevan Rajayogam and Travis Choma and Fei Liu and Aditya Barua and Colin Ji and Ji Ho Park and Vincent Hellendoorn and Alex Bailey and Taylan Bilal and Huanjie Zhou and Mehrdad Khatir and Charles Sutton and Wojciech Rzadkowski and Fiona Macintosh and Konstantin Shagin and Paul Medina and Chen Liang and Jinjing Zhou and Pararth Shah and Yingying Bi and Attila Dankovics and Shipra Banga and Sabine Lehmann and Marissa Bredesen and Zifan Lin and John Eric Hoffmann and Jonathan Lai and Raynald Chung and Kai Yang and Nihal Balani and Arthur Bražinskas and Andrei Sozanschi and Matthew Hayes and Héctor Fernández Alcalde and Peter Makarov and Will Chen and Antonio Stella and Liselotte Snijders and Michael Mandl and Ante Kärrman and Paweł Nowak and Xinyi Wu and Alex Dyck and Krishnan Vaidyanathan and Raghavender R and Jessica Mallet and Mitch Rudominer and Eric Johnston and Sushil Mittal and Akhil Udathu and Janara Christensen and Vishal Verma and Zach Irving and Andreas Santucci and Gamaleldin Elsayed and Elnaz Davoodi and Marin Georgiev and Ian Tenney and Nan Hua and Geoffrey Cideron and Edouard Leurent and Mahmoud Alnahlawi and Ionut Georgescu and Nan Wei and Ivy Zheng and Dylan Scandinaro and Heinrich Jiang and Jasper Snoek and Mukund Sundararajan and Xuezhi Wang and Zack Ontiveros and Itay Karo and Jeremy Cole and Vinu Rajashekhar and Lara Tumeh and Eyal Ben-David and Rishub Jain and Jonathan Uesato and Romina Datta and Oskar Bunyan and Shimu Wu and John Zhang and Piotr Stanczyk and Ye Zhang and David Steiner and Subhajit Naskar and Michael Azzam and Matthew Johnson and Adam Paszke and Chung-Cheng Chiu and Jaume Sanchez Elias and Afroz Mohiuddin and Faizan Muhammad and Jin Miao and Andrew Lee and Nino Vieillard and Jane Park and Jiageng Zhang and Jeff Stanway and Drew Garmon and Abhijit Karmarkar and Zhe Dong and Jong Lee and Aviral Kumar and Luowei Zhou and Jonathan Evens and William Isaac and Geoffrey Irving and Edward Loper and Michael Fink and Isha Arkatkar and Nanxin Chen and Izhak Shafran and Ivan Petrychenko and Zhe Chen and Johnson Jia and Anselm Levskaya and Zhenkai Zhu and Peter Grabowski and Yu Mao and Alberto Magni and Kaisheng Yao and Javier Snaider and Norman Casagrande and Evan Palmer and Paul Suganthan and Alfonso Castaño and Irene Giannoumis and Wooyeol Kim and Mikołaj Rybiński and Ashwin Sreevatsa and Jennifer Prendki and David Soergel and Adrian Goedeckemeyer and Willi Gierke and Mohsen Jafari and Meenu Gaba and Jeremy Wiesner and Diana Gage Wright and Yawen Wei and Harsha Vashisht and Yana Kulizhskaya and Jay Hoover and Maigo Le and Lu Li and Chimezie Iwuanyanwu and Lu Liu and Kevin Ramirez and Andrey Khorlin and Albert Cui and Tian LIN and Marcus Wu and Ricardo Aguilar and Keith Pallo and Abhishek Chakladar and Ginger Perng and Elena Allica Abellan and Mingyang Zhang and Ishita Dasgupta and Nate Kushman and Ivo Penchev and Alena Repina and Xihui Wu and Tom van der Weide and Priya Ponnapalli and Caroline Kaplan and Jiri Simsa and Shuangfeng Li and Olivier Dousse and Fan Yang and Jeff Piper and Nathan Ie and Rama Pasumarthi and Nathan Lintz and Anitha Vijayakumar and Daniel Andor and Pedro Valenzuela and Minnie Lui and Cosmin Paduraru and Daiyi Peng and Katherine Lee and Shuyuan Zhang and Somer Greene and Duc Dung Nguyen and Paula Kurylowicz and Cassidy Hardin and Lucas Dixon and Lili Janzer and Kiam Choo and Ziqiang Feng and Biao Zhang and Achintya Singhal and Dayou Du and Dan McKinnon and Natasha Antropova and Tolga Bolukbasi and Orgad Keller and David Reid and Daniel Finchelstein and Maria Abi Raad and Remi Crocker and Peter Hawkins and Robert Dadashi and Colin Gaffney and Ken Franko and Anna Bulanova and Rémi Leblond and Shirley Chung and Harry Askham and Luis C. Cobo and Kelvin Xu and Felix Fischer and Jun Xu and Christina Sorokin and Chris Alberti and Chu-Cheng Lin and Colin Evans and Alek Dimitriev and Hannah Forbes and Dylan Banarse and Zora Tung and Mark Omernick and Colton Bishop and Rachel Sterneck and Rohan Jain and Jiawei Xia and Ehsan Amid and Francesco Piccinno and Xingyu Wang and Praseem Banzal and Daniel J. Mankowitz and Alex Polozov and Victoria Krakovna and Sasha Brown and MohammadHossein Bateni and Dennis Duan and Vlad Firoiu and Meghana Thotakuri and Tom Natan and Matthieu Geist and Ser tan Girgin and Hui Li and Jiayu Ye and Ofir Roval and Reiko Tojo and Michael Kwong and James Lee-Thorp and Christopher Yew and Danila Sinopalnikov and Sabela Ramos and John Mellor and Abhishek Sharma and Kathy Wu and David Miller and Nicolas Sonnerat and Denis Vnukov and Rory Greig and Jennifer Beattie and Emily Caveness and Libin Bai and Julian Eisenschlos and Alex Korchemniy and Tomy Tsai and Mimi Jasarevic and Weize Kong and Phuong Dao and Zeyu Zheng and Frederick Liu and Fan Yang and Rui Zhu and Tian Huey Teh and Jason Sanmiya and Evgeny Gladchenko and Nejc Trdin and Daniel Toyama and Evan Rosen and Sasan Tavakkol and Linting Xue and Chen Elkind and Oliver Woodman and John Carpenter and George Papamakarios and Rupert Kemp and Sushant Kafle and Tanya Grunina and Rishika Sinha and Alice Talbert and Diane Wu and Denese Owusu-Afriyie and Cosmo Du and Chloe Thornton and Jordi Pont-Tuset and Pradyumna Narayana and Jing Li and Saaber Fatehi and John Wieting and Omar Ajmeri and Benigno Uria and Yeongil Ko and Laura Knight and Amélie Héliou and Ning Niu and Shane Gu and Chenxi Pang and Yeqing Li and Nir Levine and Ariel Stolovich and Rebeca Santamaria-Fernandez and Sonam Goenka and Wenny Yustalim and Robin Strudel and Ali Elqursh and Charlie Deck and Hyo Lee and Zonglin Li and Kyle Levin and Raphael Hoffmann and Dan Holtmann-Rice and Olivier Bachem and Sho Arora and Christy Koh and Soheil Hassas Yeganeh and Siim Põder and Mukarram Tariq and Yanhua Sun and Lucian Ionita and Mojtaba Seyedhosseini and Pouya Tafti and Zhiyu Liu and Anmol Gulati and Jasmine Liu and Xinyu Ye and Bart Chrzaszcz and Lily Wang and Nikhil Sethi and Tianrun Li and Ben Brown and Shreya Singh and Wei Fan and Aaron Parisi and Joe Stanton and Vinod Koverkathu and Christopher A. Choquette-Choo and Yunjie Li and TJ Lu and Abe Ittycheriah and Prakash Shroff and Mani Varadarajan and Sanaz Bahargam and Rob Willoughby and David Gaddy and Guillaume Desjardins and Marco Cornero and Brona Robenek and Bhavishya Mittal and Ben Albrecht and Ashish Shenoy and Fedor Moiseev and Henrik Jacobsson and Alireza Ghaffarkhah and Morgane Rivière and Alanna Walton and Clément Crepy and Alicia Parrish and Zongwei Zhou and Clement Farabet and Carey Radebaugh and Praveen Srinivasan and Claudia van der Salm and Andreas Fidjeland and Salvatore Scellato and Eri Latorre-Chimoto and Hanna Klimczak-Plucińska and David Bridson and Dario de Cesare and Tom Hudson and Piermaria Mendolicchio and Lexi Walker and Alex Morris and Matthew Mauger and Alexey Guseynov and Alison Reid and Seth Odoom and Lucia Loher and Victor Cotruta and Madhavi Yenugula and Dominik Grewe and Anastasia Petrushkina and Tom Duerig and Antonio Sanchez and Steve Yadlowsky and Amy Shen and Amir Globerson and Lynette Webb and Sahil Dua and Dong Li and Surya Bhupatiraju and Dan Hurt and Haroon Qureshi and Ananth Agarwal and Tomer Shani and Matan Eyal and Anuj Khare and Shreyas Rammohan Belle and Lei Wang and Chetan Tekur and Mihir Sanjay Kale and Jinliang Wei and Ruoxin Sang and Brennan Saeta and Tyler Liechty and Yi Sun and Yao Zhao and Stephan Lee and Pandu Nayak and Doug Fritz and Manish Reddy Vuyyuru and John Aslanides and Nidhi Vyas and Martin Wicke and Xiao Ma and Evgenii Eltyshev and Nina Martin and Hardie Cate and James Manyika and Keyvan Amiri and Yelin Kim and Xi Xiong and Kai Kang and Florian Luisier and Nilesh Tripuraneni and David Madras and Mandy Guo and Austin Waters and Oliver Wang and Joshua Ainslie and Jason Baldridge and Han Zhang and Garima Pruthi and Jakob Bauer and Feng Yang and Riham Mansour and Jason Gelman and Yang Xu and George Polovets and Ji Liu and Honglong Cai and Warren Chen and XiangHai Sheng and Emily Xue and Sherjil Ozair and Christof Angermueller and Xiaowei Li and Anoop Sinha and Weiren Wang and Julia Wiesinger and Emmanouil Koukoumidis and Yuan Tian and Anand Iyer and Madhu Gurumurthy and Mark Goldenson and Parashar Shah and MK Blake and Hongkun Yu and Anthony Urbanowicz and Jennimaria Palomaki and Chrisantha Fernando and Ken Durden and Harsh Mehta and Nikola Momchev and Elahe Rahimtoroghi and Maria Georgaki and Amit Raul and Sebastian Ruder and Morgan Redshaw and Jinhyuk Lee and Denny Zhou and Komal Jalan and Dinghua Li and Blake Hechtman and Parker Schuh and Milad Nasr and Kieran Milan and Vladimir Mikulik and Juliana Franco and Tim Green and Nam Nguyen and Joe Kelley and Aroma Mahendru and Andrea Hu and Joshua Howland and Ben Vargas and Jeffrey Hui and Kshitij Bansal and Vikram Rao and Rakesh Ghiya and Emma Wang and Ke Ye and Jean Michel Sarr and Melanie Moranski Preston and Madeleine Elish and Steve Li and Aakash Kaku and Jigar Gupta and Ice Pasupat and Da-Cheng Juan and Milan Someswar and Tejvi M. and Xinyun Chen and Aida Amini and Alex Fabrikant and Eric Chu and Xuanyi Dong and Amruta Muthal and Senaka Buthpitiya and Sarthak Jauhari and Nan Hua and Urvashi Khandelwal and Ayal Hitron and Jie Ren and Larissa Rinaldi and Shahar Drath and Avigail Dabush and Nan-Jiang Jiang and Harshal Godhia and Uli Sachs and Anthony Chen and Yicheng Fan and Hagai Taitelbaum and Hila Noga and Zhuyun Dai and James Wang and Chen Liang and Jenny Hamer and Chun-Sung Ferng and Chenel Elkind and Aviel Atias and Paulina Lee and Vít Listík and Mathias Carlen and Jan van de Kerkhof and Marcin Pikus and Krunoslav Zaher and Paul Müller and Sasha Zykova and Richard Stefanec and Vitaly Gatsko and Christoph Hirnschall and Ashwin Sethi and Xingyu Federico Xu and Chetan Ahuja and Beth Tsai and Anca Stefanoiu and Bo Feng and Keshav Dhandhania and Manish Katyal and Akshay Gupta and Atharva Parulekar and Divya Pitta and Jing Zhao and Vivaan Bhatia and Yashodha Bhavnani and Omar Alhadlaq and Xiaolin Li and Peter Danenberg and Dennis Tu and Alex Pine and Vera Filippova and Abhipso Ghosh and Ben Limonchik and Bhargava Urala and Chaitanya Krishna Lanka and Derik Clive and Yi Sun and Edward Li and Hao Wu and Kevin Hongtongsak and Ianna Li and Kalind Thakkar and Kuanysh Omarov and Kushal Majmundar and Michael Alverson and Michael Kucharski and Mohak Patel and Mudit Jain and Maksim Zabelin and Paolo Pelagatti and Rohan Kohli and Saurabh Kumar and Joseph Kim and Swetha Sankar and Vineet Shah and Lakshmi Ramachandruni and Xiangkai Zeng and Ben Bariach and Laura Weidinger and Tu Vu and Alek Andreev and Antoine He and Kevin Hui and Sheleem Kashem and Amar Subramanya and Sissie Hsiao and Demis Hassabis and Koray Kavukcuoglu and Adam Sadovsky and Quoc Le and Trevor Strohman and Yonghui Wu and Slav Petrov and Jeffrey Dean and Oriol Vinyals},
   month = {12},
   title = {Gemini: A Family of Highly Capable Multimodal Models},
   url = {http://arxiv.org/abs/2312.11805},
   year = {2023},
}

@techreport{gpt4v,
   author = {Openai},
   title = {GPT-4V(ision) System Card},
   url = {https://cdn.openai.com/papers/GPTV_System_Card.pdf},
   year = {2023},
}

@inproceedings{sundaresan2024rtsketchgoalconditionedimitationlearning,
   author = {Priya Sundaresan and Quan Vuong and Jiayuan Gu and Peng Xu and Ted Xiao and Sean Kirmani and Tianhe Yu and Michael Stark and Ajinkya Jain and Karol Hausman and Dorsa Sadigh and Jeannette Bohg and Stefan Schaal},
   booktitle = {RSS 2024 Workshop: Data Generation for Robotics},
   month = {3},
   title = {RT-Sketch: Goal-Conditioned Imitation Learning from Hand-Drawn Sketches},
   url = {http://arxiv.org/abs/2403.02709},
   year = {2024},
}

@inproceedings{gu2023rttrajectory,
   author = {Jiayuan Gu and Sean Kirmani and Paul Wohlhart and Yao Lu and Montserrat Gonzalez Arenas and Kanishka Rao and Wenhao Yu and Chuyuan Fu and Keerthana Gopalakrishnan and Zhuo Xu and Priya Sundaresan and Peng Xu and Hao Su and Karol Hausman and Chelsea Finn and Quan Vuong and Ted Xiao},
   booktitle = {The Twelfth International Conference on Learning Representations},
   title = {RT-Trajectory: Robotic Task Generalization via Hindsight Trajectory Sketches},
   url = {https://openreview.net/forum?id=F1TKzG8LJO},
   year = {2024},
}

@inproceedings{draw-map,
   author = {Federico Boniardi and Bahram Behzadian and Wolfram Burgard and Gian Diego Tipaldi},
   doi = {10.1109/ECMR.2015.7324188},
   isbn = {978-1-4673-9163-4},
   booktitle = {2015 European Conference on Mobile Robots (ECMR)},
   month = {9},
   pages = {1-6},
   publisher = {IEEE},
   title = {Robot navigation in hand-drawn sketched maps},
   year = {2015},
}
\end{document}